\documentclass[letterpaper, 10 pt, conference]{ieeeconf}  

\IEEEoverridecommandlockouts                              

\usepackage{subcaption}
\usepackage{adjustbox}
\usepackage{placeins}

\usepackage{graphicx} 
\usepackage{caption}
\usepackage{booktabs}
\usepackage{tabularx}
\usepackage{makecell} 

\newcommand{\NA}{\textemdash}  

\usepackage{url}               
\usepackage{amsmath} 
\usepackage{amssymb}  
\usepackage{comment}
\usepackage{tcolorbox}
\usepackage[noadjust]{cite}       
\usepackage[hidelinks]{hyperref}

\title{\LARGE \bf
Robustness Is a Function, Not a Number: A Factorized Comprehensive Study of OOD Robustness in Vision-Based Driving
}

\definecolor{authors}{RGB}{50, 50, 180}
\author{
    \textcolor{authors}{Amir Mallak}$^{1}$ \quad
    \textcolor{authors}{Alaa Maalouf}$^{1}$  \\
    {\normalsize $^1$\textcolor{magenta}{University of Haifa}} \\
    {\tt\small Correspondance: mallak002@gmail.com}
}

\begin{document}

\maketitle

\thispagestyle{empty}
\pagestyle{empty}


\begin{abstract}
Out-of-distribution (OOD) robustness in autonomous driving is often reduced to a single number, hiding what breaks a policy. We decompose environments along five axes: scene (rural/urban), season, weather, time (day/night), and agent mix; and measure performance under controlled \(k\)-factor perturbations \((k\in\{0,1,2,3\})\). Using closed-loop control in VISTA, we benchmark FC, CNN, and ViT policies, train compact ViT heads on frozen foundation-model (FM) features, and vary ID support in scale, diversity, and temporal context. (1) ViT policies are markedly more OOD-robust than comparably sized CNN/FC, and FM features yield state-of-the-art success at a latency cost. (2) Naive temporal inputs (multi-frame) do not beat the best single-frame baseline. (3) The largest single-factor drops are rural \(\rightarrow\) urban and day \(\rightarrow\) night (\(\sim31\%\) each); actor swaps \(\sim10\%\), moderate rain \(\sim7\%\); season shifts can be drastic, and combining a time flip with other changes further degrades performance. (4) FM-feature policies stay above \(85\%\) under three simultaneous changes; non-FM single-frame policies take a large first-shift hit, and all no-FM models fall below \(50\%\) by three changes. (5) Interactions are non-additive: some pairings partially offset, whereas season-time combinations are especially harmful. (6) Training on winter/snow is most robust to single-factor shifts, while a rural+summer baseline gives the best overall OOD performance. (7) Scaling traces/views improves robustness (\(+11.8\) points from 5 to 14 traces), yet targeted exposure to hard conditions can substitute for scale. (8) Using multiple ID environments broadens coverage and strengthens weak cases (urban OOD \(60.6\%\rightarrow70.1\%\)) with a small ID drop; single-ID preserves peak performance but in a narrow domain. These results yield actionable design rules for OOD-robust driving policies.
\end{abstract}

\section{Introduction}

Autonomous driving systems must operate far beyond the narrow slice of conditions seen during training. Real roads combine many shifting \emph{factors}: scene layout (rural vs.\ urban), time of day, season, weather, and the mix of nearby agents (vehicles, pedestrians, animals). Small changes along any one axis can subtly alter visual appearance, dynamics, and affordances; combined changes amplify these effects. Despite rapid progress in perception and end-to-end control, a core open question remains: \emph{which factors matter most for out-of-distribution (OOD) robustness, and how should we design the in-distribution (ID) training pipeline support to prepare for them?}

\textbf{To answer this,} we advocate a \emph{factorized} view of distribution shift. Rather than treating “OOD” as a monolith, we explicitly decompose the environment into semantically meaningful axes and evaluate policies under controlled $k$-factor perturbations—i.e., test conditions that differ from the in-distribution (ID) support on exactly $k$ factors. 
This yields interpretable robustness profiles: performance as a function of \emph{how many} factors change and \emph{which} factors change. Our analysis shows that such decomposition exposes sensitivities that are obscured by aggregate OOD scores.

\textbf{Why this matters?} 
Safety-critical deployment hinges on reliable behavior under inevitable distribution shift, e.g., night drives after a model trained at noon, first snow of the season after a summer-only dataset, or an unexpected animal entering the roadway. A factorized evaluation makes robustness \emph{diagnosable}: practitioners can identify, for example, that weather+night degrades steering more than scene+agents, or that balancing time-of-day in the ID set yields larger gains than balancing season, given a fixed data budget. Such insights directly inform data collection, simulation curriculum design, and model selection.

\subsection{Our contributions}
Motivated by the discussion above, we present the first \emph{systematic, factorized} experimental study of generalization in vision-based autonomous driving. We quantify how (i) the \emph{training-data factors} included in the ID set, (ii) the \emph{type and number} of test-time distribution shifts, (iii) the \emph{policy architecture} (MLP, CNN, ViT) and the use of \emph{foundation-model features}, and (iv) key \emph{design choices} (data budget, ID diversity vs.\ scale, single-frame vs.\ sequence) impact OOD robustness. Specifically, we contribute:

\begin{itemize}
\item \textbf{A factorized OOD framework.} We formalize the environment as a Cartesian product of factor sets and define $k$-factor OOD shells via a Hamming distance over factors. This provides a precise and reproducible way to construct ID/OOD splits and to attribute errors to specific axes of variation.
 \item \textbf{Systematic architectural comparison.} Under matched training budgets and protocols, we benchmark FC, CNN, and ViT policies on closed-loop metrics and regression error, reporting robustness as a function of the number and identity of shifted factors.
\item \textbf{What to include in the ID set.} We vary the ID support along selected factors \emph{at constant data budget, and raising budget in terms of diversity and quantity of the same ID} to quantify which axes are most valuable for OOD generalization and when broad coverage trades off with ID specialization.
\item \textbf{Foundation-model features for control.} Using frozen DINO/BLIP-2 patch descriptors with a compact ViT policy head, we isolate the contribution of generic visual features to OOD robustness and analyze how these benefits interact with the choice of ID support.
 \item \textbf{Temporal context.} We compare single-frame policies to sequence-based models to assess whether short histories mitigate specific factor shifts (e.g., adverse weather) and how temporal aggregation complements foundation-model features.
\end{itemize}

\section{Related Work}
\textbf{From modular stacks to end-to-end policies}
Classical systems used a modular stack (perception→prediction→planning→control) that was reliable yet prone to compounding errors. End-to-end control dates to \textsc{ALVINN}~\cite{Pomerleau1989} and has since advanced to pixels to steering and learned affordances~\cite{Bojarski2016,Chen2015DeepDriving,amini2019variational,wang2023learning,xiao2023barriernet}. Conditional imitation learning adds high-level commands~\cite{Codevilla2018}, while later analyses expose limits of pure behavior cloning~\cite{Codevilla2019}. We retain the end-to-end setting and ask which architectural biases (MLP/CNN/ViT) and which training distributions best support robustness under controlled shifts.

\textbf{Generalization and robustness under distribution shift. }
OOD sensitivity—across towns, weather, lighting, and traffic—has been documented repeatedly; for example, performance drops starkly in new towns or adverse weather even when ID results look strong~\cite{Codevilla2019}.
Common remedies include domain randomization and augmentation~\cite{Tobin2017}, and domain adaptation; yet open-loop gains often fail to translate to closed-loop safety. We complement these lines by \emph{factorizing} shift along semantically meaningful axes (scene, time, season, weather, agents) and measuring robustness as a function of \emph{how many} and \emph{which} factors change.

\textbf{Foundation models for vision and their use in driving}
Large-scale pretraining yields image encoders whose features transfer broadly: CLIP aligns images with language for robust zero-shot recognition~\cite{Radford2021CLIP}, while DINO learns strong self-supervised ViT representations with emergent semantics~\cite{Caron2021DINO}; BLIP-2 efficiently couples frozen vision encoders to LLMs~\cite{Li2023BLIP2}. While leveraging LLMs in a zero shot manner for driving has proven to be week~\cite{sreeram2025probing}, driving-specific pretraining has leveraged diverse web or fleet data for policy representations~\cite{Zhang2022YouTube,Chen2022LAV,Wang2024Drive,mallak2026see} and showed to be robust across a variety of tasks~\cite{maalouf2023follow,chahine2024flexendtoendtextinstructedvisual,chahine2025decentralized}. We operationalize this idea in control by feeding \emph{frozen, patch-wise} features (DINO/CLIP/BLIP-2) to a compact policy head and quantifying when such features improve OOD robustness—and along which factors.

\textbf{Structured, factorized evaluation}
Simulation enables controlled manipulations of environment factors.
CARLA~\cite{Dosovitskiy2017CARLA} popularized \emph{New Town} and \emph{New Weather} splits; NoCrash~\cite{Codevilla2019} contrasted traffic density and weather to expose failure modes.
Data-driven simulators like VISTA~\cite{Amini2022VISTA2} reproject real logs to photorealistic, closed-loop scenes, supporting reproducible stress tests. We formalize factorization by defining $k$-factor OOD \emph{shells} via a Hamming distance over factors, enabling matched-budget, per-axis attribution rather than a single aggregate OOD.

\textbf{Temporal modeling for control.}
Temporal context improves driving decisions over single-frame policies.
Early FCN--LSTM models fused video history for egomotion prediction~\cite{Xu2017FCNLSTM}, and recent end-to-end approaches use spatial--temporal Transformers for perception, prediction, planning~\cite{Hu2022STP3} or explicit temporal/global reasoning~\cite{Shao2023ReasonNet}. We directly compare single-frame and sequence-based policies (temporal ViT and RNN heads) under our factorized OOD protocol to reveal which shifts benefit most from temporal aggregation.

\textbf{Benchmarks and surveys.}
A growing literature benchmarks end-to-end stacks and catalogs open challenges in robustness, causality, and evaluation~\cite{Zhu2021SurveyAD,Chen2023E2ESurvey}.
Simulators like CARLA and VISTA remain central for closed-loop, controllable, and reproducible experiments~\cite{Dosovitskiy2017CARLA,Amini2022VISTA2}. Our work contributes a \emph{methodological} addition—factorized OOD shells with matched-budget comparisons across architectures, training supports, and temporal context—intended to complement existing benchmarks and inform data curation for real-world deployment.

\section{Experimental Setup}
We first present the exact questions we aim to study.

\subsection{Key questions we address}
This work answers the following practical questions:
\begin{itemize}
 \item \textbf{Q1 — Architecture vs.\ robustness.} Under matched training protocols, which backbone (FC, CNN, ViT) is intrinsically more resilient to specific factor shifts?
    \item \textbf{Q2 — Role of foundation-model features.} Do frozen patch-wise features from DINO/BLIP-2 confer uniform robustness, or do they target specific axes (e.g., lighting) while being neutral elsewhere?
    \item \textbf{Q3 — Temporal context.} Do short frame histories improve robustness (and for which factors), compared to single-frame policies?
    \item \textbf{Q4 — Which factors matter most?} Among scene, time, season, weather, and agent mix, which single-factor shifts degrade performance the most?
    \item \textbf{Q5 — How many changes can a policy tolerate?} How does performance decay as the number of shifted factors \(k\) increases (1, 2, 3), and is the decay monotonic?
    \item \textbf{Q6 — Are factor interactions additive?} Do combinations such as night+snow hurt more than the each of their parts (super-additivity), or less?
   
    \item \textbf{Q7 — Training data choices.} Under which settings is it better to train a model so that it can generalize to unseen configurations? For example, is it more effective to train on night or day data, in summer or winter conditions, or in rural versus urban environments?
    \item \textbf{Q8 — Data diversity.} Does increasing ID \emph{diversity} (more factor coverage) help OOD generalization? 
    \item \textbf{Q9 — Data Scale.} Does increasing \emph{quantity} of a narrow ID help OOD generation?
    \item \textbf{Q10 — Specialization vs.\ Generalization.} How does increasing the diversity of the ID data trade off specialization (performance on a targeted ID subset) against generalization (robustness to OOD shifts)?    
\end{itemize}

\subsection{Task Formulation and Platform}
We study closed-loop control for autonomous driving in the \emph{VISTA} simulator~\cite{amini2022vista}, a photorealistic, data-driven environment designed for learning-based autonomy. At simulation step $t$, the agent receives either a single RGB frame $I_t \in \mathbb{R}^{H \times W \times 3}$ or a short sequence of $\tau$ frames $I_{t-\tau:t}$, and must predict continuous controls $\hat{\theta}_t$ (steering angle) and $\hat{g}_t$ (throttle/gas). The policy hence realizes a mapping
$
\pi: I_{t-\tau:t} \mapsto \big(\hat{\theta}_t, \hat{g}_t\big).
$ 
Unless otherwise stated, we use single-frame input ($\tau{=}0$); the multi-frame setting is introduced in Sec.~\ref{subsec:temporal}.

\subsection{Environment Factorization and Distribution Shifts}
\label{subsec:factorization}
To make distribution shifts precise, we factorize the environment along five semantically meaningful axes that strongly affect driving:
\begin{itemize}
    \item \textbf{Scene type:} $\mathcal{S}=\{\text{rural},\text{urban}\}$
    \item \textbf{Season:} $\mathcal{Se}=\{\text{summer},\text{winter},\text{spring}, \text{fall}\}$
    \item \textbf{Weather:} $\mathcal{W}=\{\text{dry},\text{rain},\text{snow}\}$
    \item \textbf{Time of day:} $\mathcal{T}=\{\text{day},\text{night}\}$
    \item \textbf{Agents/characters:} traffic actors and non-vehicle entities (e.g.,$\mathcal{A}=\{\text{cars},\text{animals}\ \text{(etc.)}\}$).
\end{itemize}

The full environment configuration space is the Cartesian product of the above factor levels
\[
\boxed{\ \mathcal{E}\;=\;\mathcal{S}\times\mathcal{T}\times\mathcal{Se}\times\mathcal{W}\times\mathcal{A}\ }, 
\]
and we denote a specific environment configuration by a tuple $e=(s,t,\,\sigma,\,w,\,a)\in \mathcal{E}$.Thus, the \emph{in-distribution} (ID) training set is supported on a designated subset $\mathcal{E}_{\text{ID}}\subseteq \mathcal{E}$. 

\subsection{Levels of OOD shifts} 
A \emph{$k$-factor OOD} test condition, written $e'\in\mathcal{E}^{(k)}_{\text{OOD}}$, differs from the ID support on exactly $k$ factors (with all other factors matched). We evaluate across $k\in\{0, 1,2,3\}$ and report results per-factor (which factor changed) and per-$k$ (how many factors changed). 
For each study, we: (i) specify $\mathcal{E}_{\text{ID}}$; (ii) construct test suites for all admissible $k$-factor changes that are feasible in VISTA; and (iii) ensure no scene instances from $\mathcal{E}_{\text{ID}}$ appear in OOD tests. When $\mathcal{E}_{\text{ID}}$ is a mixture (Sec.~\ref{subsec:train-mixtures}), we stratify by factor to avoid inadvertent leakage.

\subsection{Basic Policy Models and Training}
\label{subsec:models}
We compare three vision policy backbones trained end-to-end from images to controls: (1) \textbf{FC (Fully-Connected):} a shallow MLP operating on spatially downsampled/flattened pixels. (2) \textbf{CNN:} a standard convolutional network with global pooling and a control head. (3) \textbf{ViT:} a Vision Transformer with patch embedding and a control head.
All models output $(\hat{\theta},\hat{g})$ each step. We optimize the weighted regression Mean Squared Error (MSE) loss
\[
\mathcal{L} = \lambda_{\theta}\,\mathrm{MSE}(\theta,\hat{\theta}) + \lambda_{g}\,\mathrm{MSE}(g,\hat{g}),
\]
with $\lambda_{\theta},\lambda_{g}$ fixed across studies. Unless otherwise noted, image encoders are frozen, and the policy model is trained from scratch on the specified ID distribution.

\subsection{Foundation-model Feature Policies (Sec.~\ref{subsec:fm}).}
For the ViT policy head, we also consider policies that consume \emph{frozen} patch-wise features from large-scale pre-trained encoders (DINO and BLIP-2). Given a frame $I_t$, we extract per-patch descriptors $\{z_{t,p}\}$ (as explained in~\cite{amir2021deep,jatavallabhula2023conceptfusion} for DINO, and in~\cite{Wang2024Drive,chahine2024flexendtoendtextinstructedvisual} for BLIP2) and feed them as tokens into the policy backbone (a compact ViT), which is trained to predict controls while the feature extractor is frozen. This isolates the effect of generic, Internet-scale features on OOD robustness.

\subsection{Evaluation Metrics and Protocol}
\label{subsec:metrics}
We report the closed-loop performance by evaluating \textbf{Route completion} (\%) and \textbf{Infraction counts} (collisions, off-road, lane departures) under closed-loop roll-outs.
For each configuration, we average over multiple seeds and routes, reporting mean~$\pm$~std. Statistical comparisons use paired tests across matched seeds/routes with Holm correction. Training/validation/test splits are fixed per study; hyperparameters are shared across models unless an ablation explicitly changes them. Additional protocol details, including horizon, episode counts, and statistical testing, are provided in Appendix~\ref{app:metrics_details}.

\subsection{Study S1: Architecture Robustness and OOD Factorized Shifts}
\label{subsec:s1}
We train FC, CNN, and ViT on a fixed ID distribution $\mathcal{E}_{\text{ID}}$ and evaluate on OOD sets with $k\in\{1,2,3\}$ factor changes. We analyze: (1) sensitivity curves as a function of $k$ (\emph{how many} factors change), and how it affects each policy model, (2) per-factor robustness: \emph{which} factor changes affect each policy the most (the least), (3) interactions between factors (e.g., night+snow vs.\ night+rain) and their effect, and inherently, and (4) architecture robustness to all of these factors and level of changes. 
This isolates architectural inductive biases with identical data budgets and training schedules.

\subsection{Study S2: Effect of the ID Training Distribution}
\label{subsec:s2}
We repeat S1 while redefining $\mathcal{E}_{\text{ID}}$ to examine how the choice of training factors influences OOD generalization. Specifically, we consider two alternative ID configurations from $\mathcal{E}$. In addition, to study the effect of training data size under a fixed configuration, we train the same ViT model on that configuration using 1, 5, and 14 traces.

We quantify how the choice of ID factors impacts robustness profiles from S1 for each architecture, levels and factors of change, by repostring the metric defined in Section~\ref{subsec:metrics}.

\begin{figure*}[t]
    \centering
\includegraphics[width=1.0\textwidth]{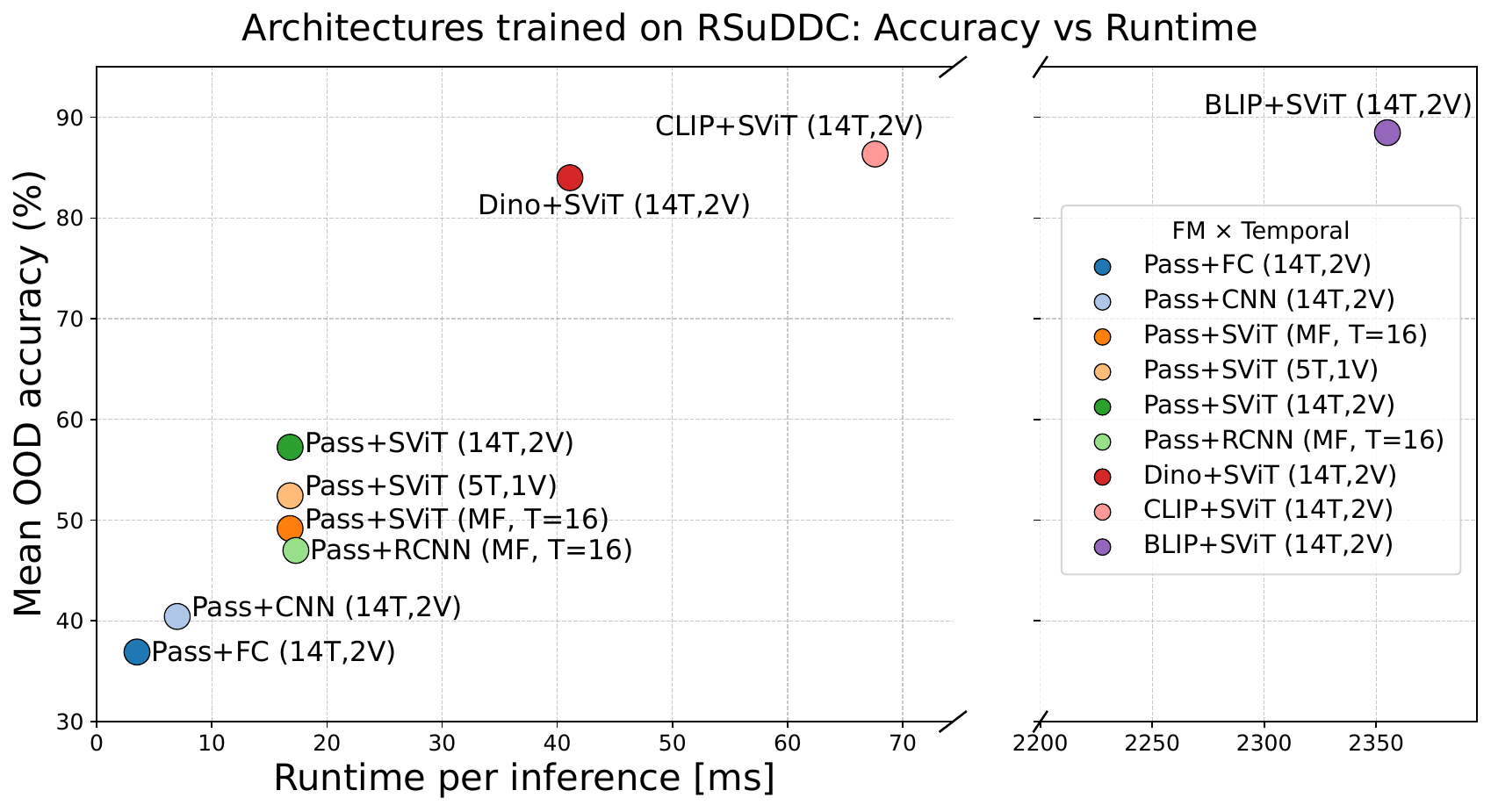}
\caption{Accuracy as a function of runtime across model variants.}
\label{fig:arch_accuracy_vs_runtime}
\end{figure*}

\subsection{Study S3: Foundation-Model Features with the Best Backbone}
\label{subsec:fm}
From S1 we select the strongest backbone (empirically ViT) and repeat S1 using frozen patch-wise features from {DINO}, {Clip}, and {BLIP-2}. We train compact policy heads (ViT-based) on top of features and evaluate $k$-factor OOD shifts. We ask: (1)  Do generic features improve OOD robustness relative to training from scratch? (2) Are gains uniform across factors (e.g., weather vs.\ time-of-day)?
    (3) How do FM-feature policies interact with the ID choice from S2? 
    (4) Which foundation model helps generalize better? on what factors? and what level of change?

\subsection{Study S4: Data Scale and Diversity vs.\ Specialization}
\label{subsec:train-mixtures}
We vary both \emph{quantity} and \emph{diversity} of ID data and compare:
\begin{enumerate}
    \item \textbf{Single-ID$\rightarrow$Same-ID:} train and test on the same single configuration (upper bound on specialization).
    \item \textbf{Single-ID$\rightarrow$Other-ID:} train on one configuration, test on a different single configuration (pure shift).
    \item \textbf{Multi-ID$\rightarrow$Single-ID:} train on a mixture of configurations, test on a single target configuration (does diversity harm specialization?).
\end{enumerate}
Each condition is evaluated for $k\in\{0,1,2,3\}$ factor changes, with and without FM features (from S3). This study disentangles the effect of more examples and \emph{varied} examples in the ID set to in and out of distributions.

\subsection{Study S5: Temporal Context---Single Frame vs.\ Sequence}
\label{subsec:temporal}
We compare \emph{single-frame} policies ($\tau{=}0$) against \emph{multi-frame} policies using short histories ($\tau{>}0$). For the temporal models we evaluate: (1) \textbf{ViT-Temporal:} a ViT backbone operating on per-frame tokens with a lightweight temporal aggregator (e.g., temporal pooling or attention across frames).
    (2) \textbf{RCNN-Temporal:} a CNN encoder with a recurrent head over frame-level embeddings.  We re-run S1, S2, and S4 under both temporal settings to quantify how motion cues affect OOD robustness and whether temporal context complements FM features.

\subsection{Implementation Details}
\label{subsec:impl}
We use AdamW with cosine decay, standard data normalization, and early stopping on validation MSE. Closed-loop evaluations follow the same seeds and route sets across models. Full training and evaluation configuration details, including temporal clip settings and compute setup, are provided in Appendix~\ref{app:impl_details}.

\section{Results}
In this section, we present our results from a top-level perspective: we begin with a broad summary of the findings and then delve into the detailed analysis.

\noindent\textbf{Terminology.}
In the following figures and discussion, we adopt the naming convention for each trained model
\[
\langle \text{FeatureExtractor} \rangle + \langle \text{PolicyArchitecture} \rangle \; (n_{\text{train}}, n_{\text{val}}),
\]
where $n_{\text{train}}$ and $n_{\text{val}}$ denote the number of traces used for training and validation, respectively. 
The term $\langle \text{FeatureExtractor} \rangle$ refers to the foundation model (FM) backbone used to extract features (DINO/CLIP/BLIP2). If no feature extractor is used, we write 
\texttt{Pass+<PolicyArchitecture>}. The term $\langle \text{PolicyArchitecture} \rangle$ refers to the policy model (e.g., CNN, fully connected (FC), or Vision Transformer (ViT)). 
Whenever multiple frames are used in training, we append \texttt{T=16} to specify training on 16 consecutive frames.
\textbf{Environment naming convention.} To concisely encode environment conditions, we use 5\textbackslash6-character tags (e.g., \texttt{RFDNC}, \texttt{RSuDDC}, \texttt{USuRDC}), where each position denotes a specific semantic factor: \textbf{Scene}, \textbf{Season}, \textbf{Weather}, \textbf{Time}, and \textbf{Actor}. The first character encodes \textbf{Scene} as \texttt{R} (rural) or \texttt{U} (urban); the second\textbackslash third, \textbf{Season} as \texttt{Su} (Summer) or \texttt{Sp} (Spring) or \texttt{W} (winter) or \texttt{F} (Fall); the third\textbackslash forth, \textbf{Weather} as \texttt{D} (dry), \texttt{R} (Rain), or \texttt{S} (Snow); the fourth\textbackslash fifth, \textbf{Time} as \texttt{D} (day) or \texttt{N} (night); and the fifth\textbackslash sixth, \textbf{Actor} as \texttt{C} (Car) or \texttt{A} (Animal). As an example, \texttt{RSuDDC} refers to a Rural-Summer-Dry-Day-Car; \texttt{UFRNA} denotes Urban-Fall-Rain-Night-Animal. This notation enables compact, interpretable reference to specific environmental configurations throughout the paper.

\subsection{Architectures and training choices}

\begin{figure}[!t]
  \centering

  \subfloat[Single factor\label{fig:EOC_Stage1}]{
    \adjustbox{valign=t}{
      \includegraphics[width=\columnwidth,height=0.27\textheight,keepaspectratio,
      trim=0.5pt 0.5pt 0.5pt 0.5pt,clip]{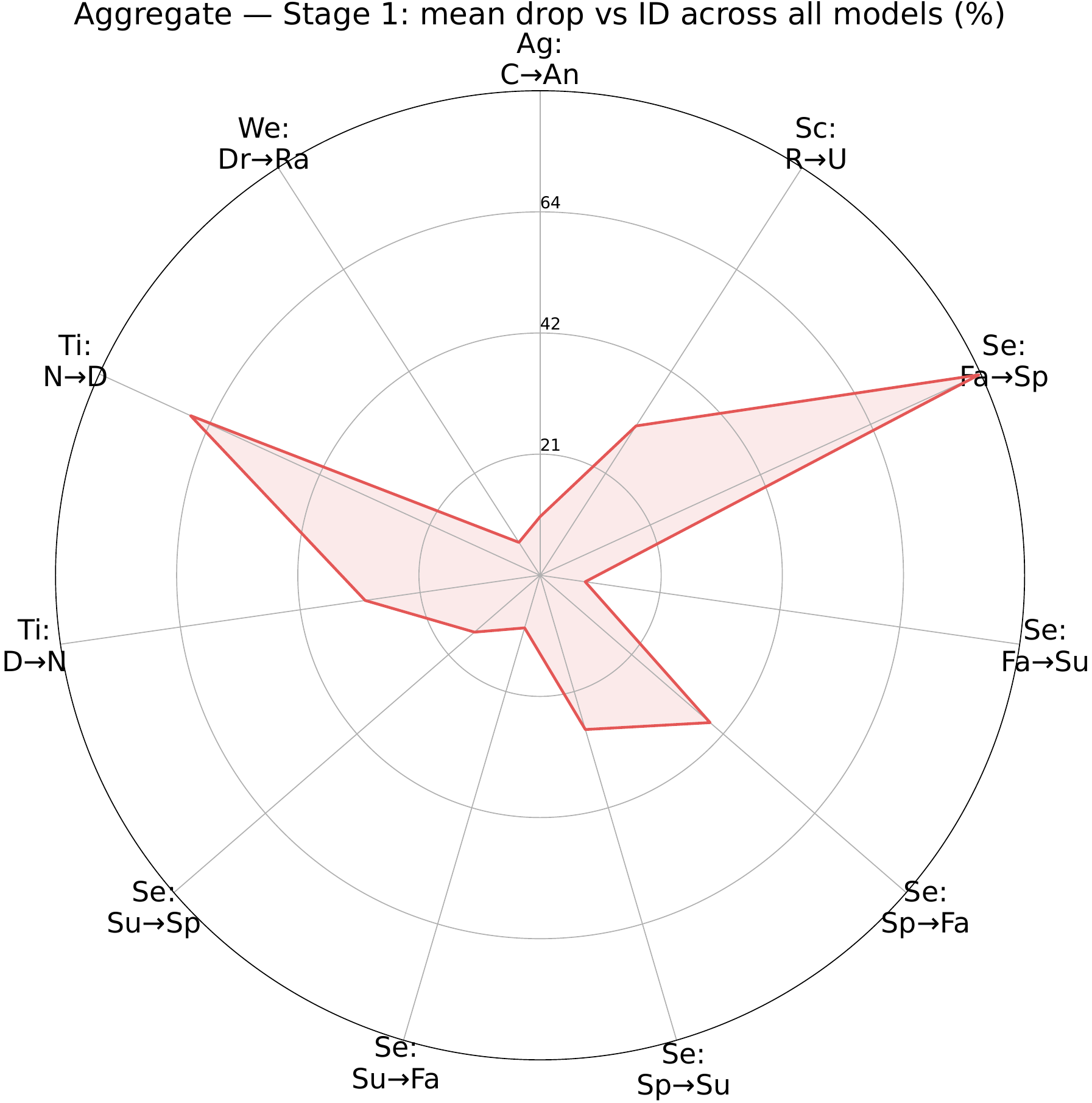}
    }
  }

  \vspace{0.35em}

  \subfloat[Double factor\label{fig:EOC_Stage2}]{
    \adjustbox{valign=t}{
      \includegraphics[width=\columnwidth,height=0.27\textheight,keepaspectratio,
      trim=0.5pt 0.5pt 0.5pt 0.5pt,clip]{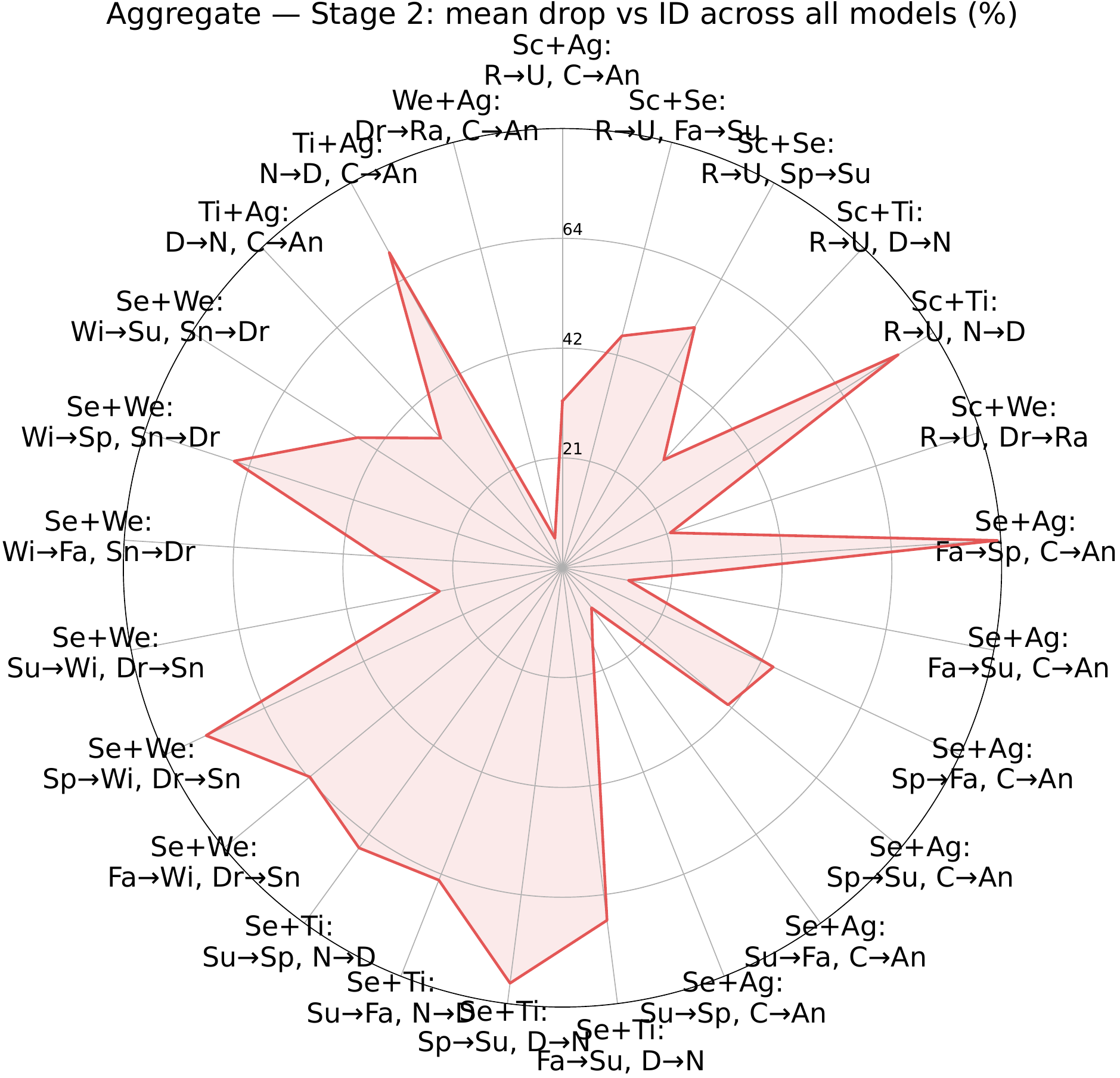}
    }
  }

  \vspace{0.35em}

  \subfloat[Triple factor\label{fig:EOC_Stage3}]{
    \adjustbox{valign=t}{
      \includegraphics[width=\columnwidth,height=0.27\textheight,keepaspectratio,
      trim=0.5pt 0.5pt 0.5pt 0.5pt,clip]{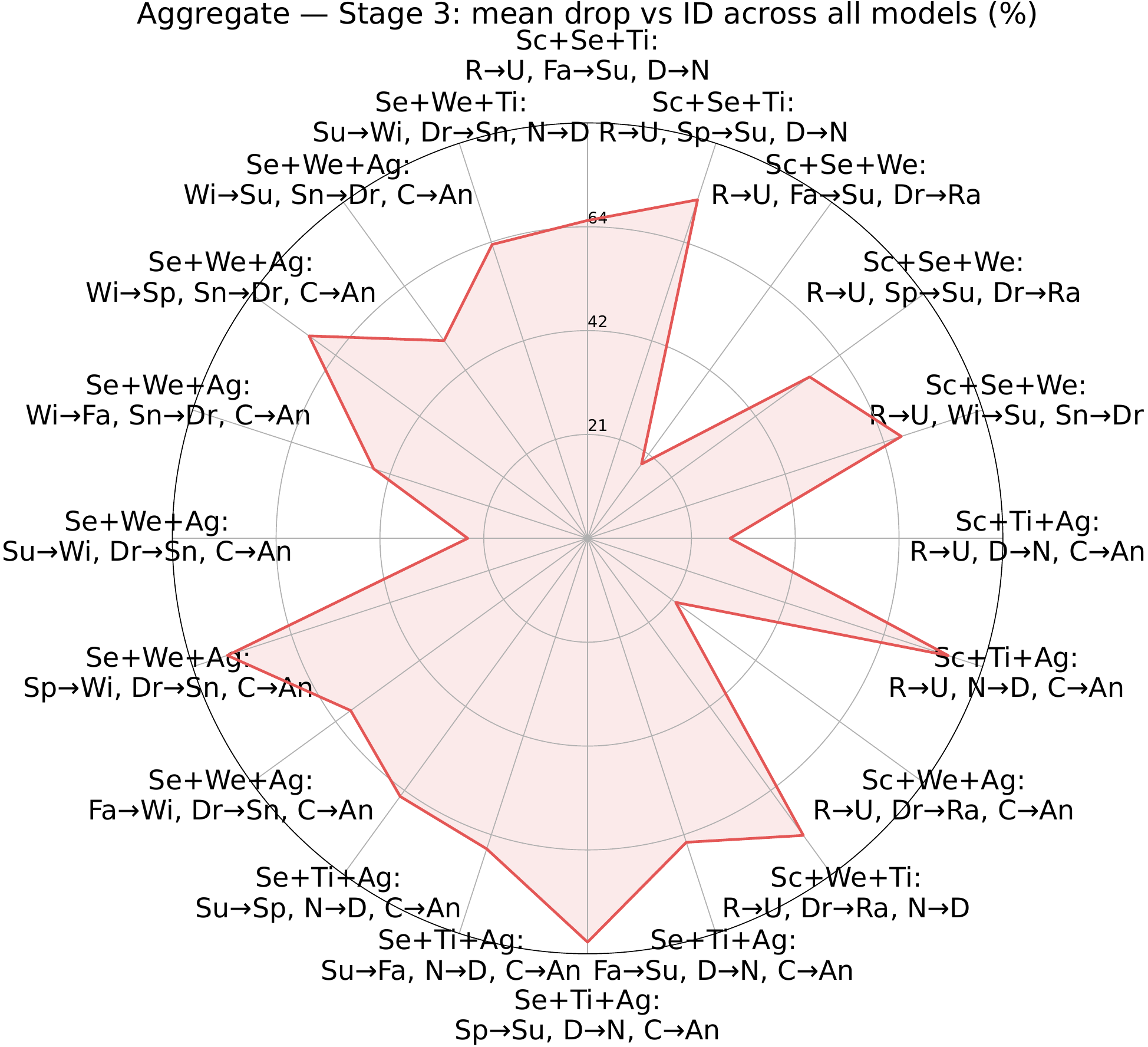}
    }
  }

  \caption{
  Effect of changes across one, two, and three simultaneous factors.
  \textbf{Key:} Sc, scene (R, rural; U, urban),
  Se, season (Wi, winter; Sp, spring; Su, summer; Fa, fall),
  We, weather (Dr, dry; Ra, rain; Sn, snow),
  Ti, time (D, day; N, night),
  Ag, agents (C, car; An, animal).
  }
  \label{fig:EOC_all}
\end{figure}

\textbf{Architecture vs. robustness.}
For context, we trained CNN, FC, and a simple ViT on the same in-distribution setting (Rural–Summer–Dry–Day–Car) using 14 traces without any feature extractor.  
In addition, we trained another ViT on only 5 traces from the same configuration to evaluate performance under limited data availability.
We then trained a ViT with each of the three feature extractors (DINO/CLIP/BLIP2), and finally, a set of models trained on the last 16 frames as input 
(see Section~\ref{subsec:temporal} for details).

Fig.~\ref{fig:arch_accuracy_vs_runtime} summarizes runtime per inference vs.\ mean closed-loop success across OOD scenarios.
With a fixed (non-FM) extractor, upgrading the policy architecture markedly improves robustness: Pass+FC and Pass+CNN reach $36.9\%$ and $40.4\%$, whereas Pass+ViT attains $57.2\%$ at $\sim 16.8$\,ms ($\approx\!+16.8$ points over CNN for $\approx\!2.4\times$ latency). This indicates that:

\begin{tcolorbox}[colback=gray!10,colframe=black!50,title=Takeaway 1,
boxsep=2pt, left=3pt, right=3pt, top=5pt, bottom=5pt]
Even without external pretraining,\textbf{ViT is a stronger policy and substantially boosts out-of-distribution performance compared to a CNN of the same size.}
\end{tcolorbox}

\textbf{Role of foundation-model features.} 
Replacing the no-FM extractor with FM features yields a large jump in accuracy: Dino+ViT and CLIP+ViT achieve $84.0\%$ and $86.4\%$, respectively, while BLIP2+ViT reaches $88.5\%$.
These gains come with higher cost: relative to Pass+ViT ($\sim 16.8$\,ms), Dino requires $\sim 2.4\times$ runtime, CLIP $\sim 4\times$, and BLIP2 is over two orders of magnitude slower ($\sim 2355$\,ms). Thus

\begin{tcolorbox}[colback=gray!10,colframe=black!50,title=Takeaway 2,
boxsep=2pt, left=3pt, right=3pt, top=5pt, bottom=5pt]
\textbf{Foundation models features offer state-of-the-art robustness but at significant latency, which may limit real-time deployment.}
\end{tcolorbox}

\begin{figure*}[!t]
\centering

\begin{subfigure}[t]{0.32\textwidth}
  \centering
  \includegraphics[height=0.19\textheight,keepaspectratio]{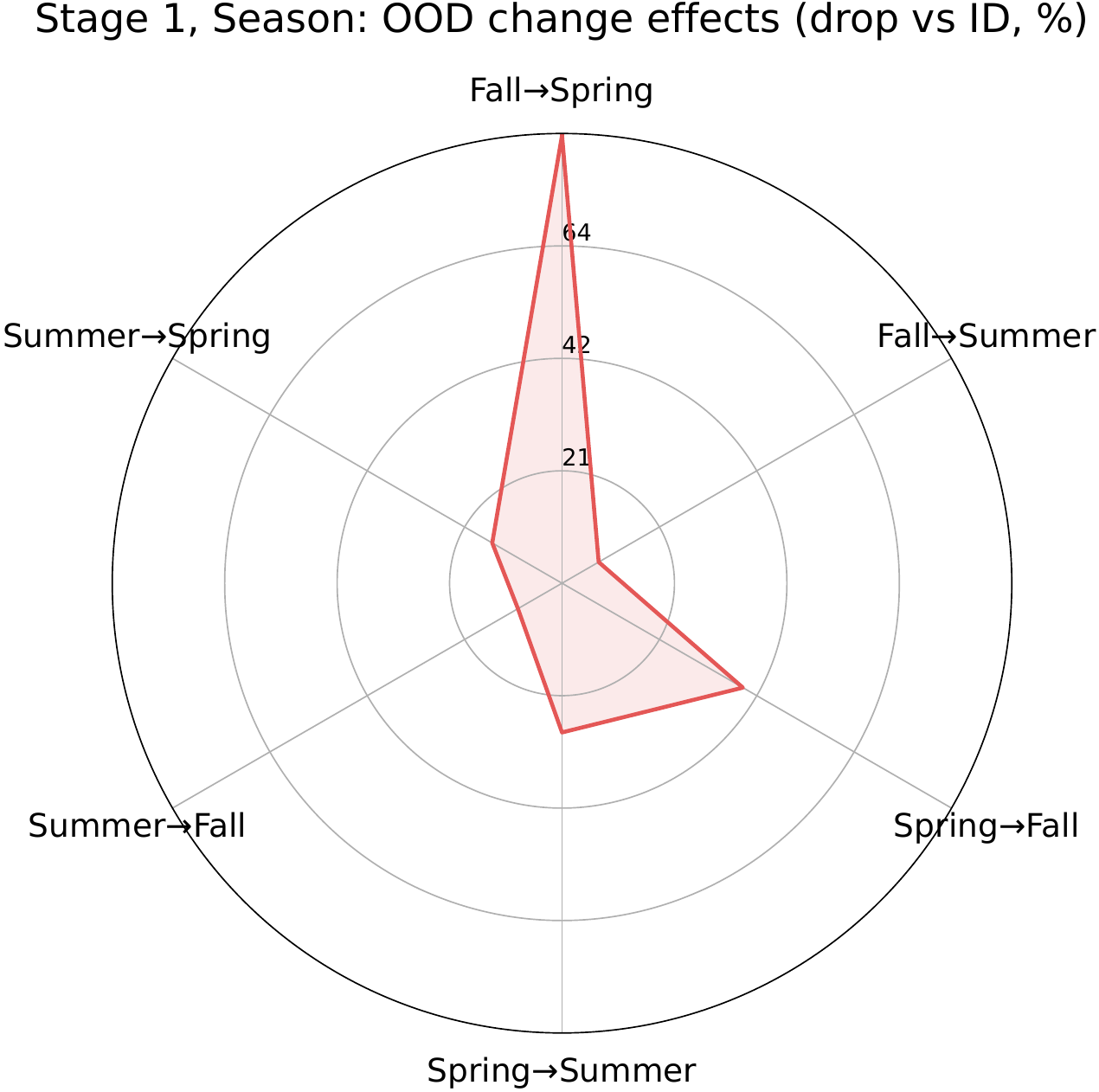}
  \caption{Stage 1, Season}
\end{subfigure}\hfill
\begin{subfigure}[t]{0.32\textwidth}
  \centering
  \includegraphics[height=0.19\textheight,keepaspectratio]{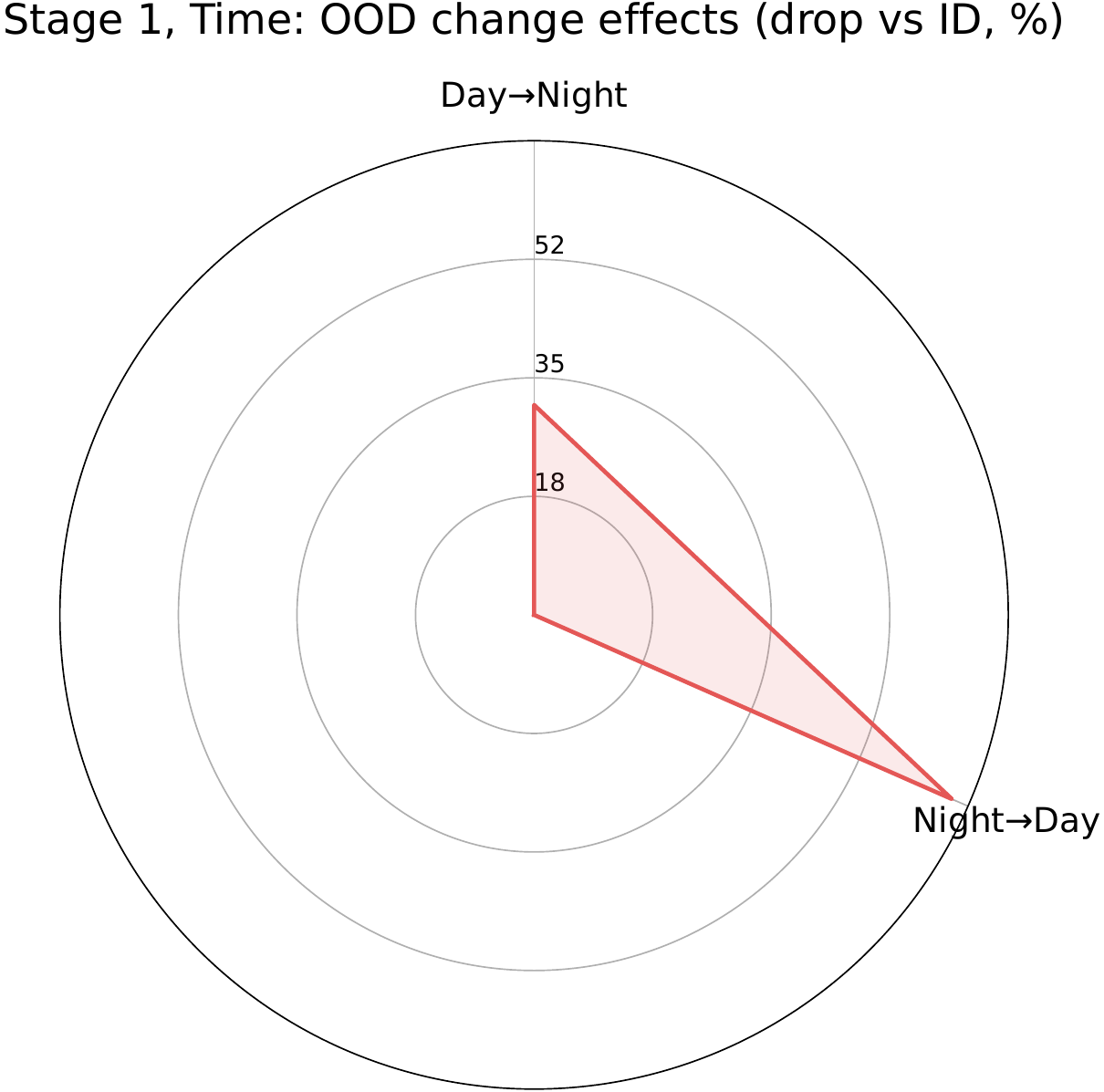}
  \caption{Stage 1, Time}
\end{subfigure}\hfill
\begin{subfigure}[t]{0.32\textwidth}
  \centering
  \includegraphics[height=0.19\textheight,keepaspectratio]{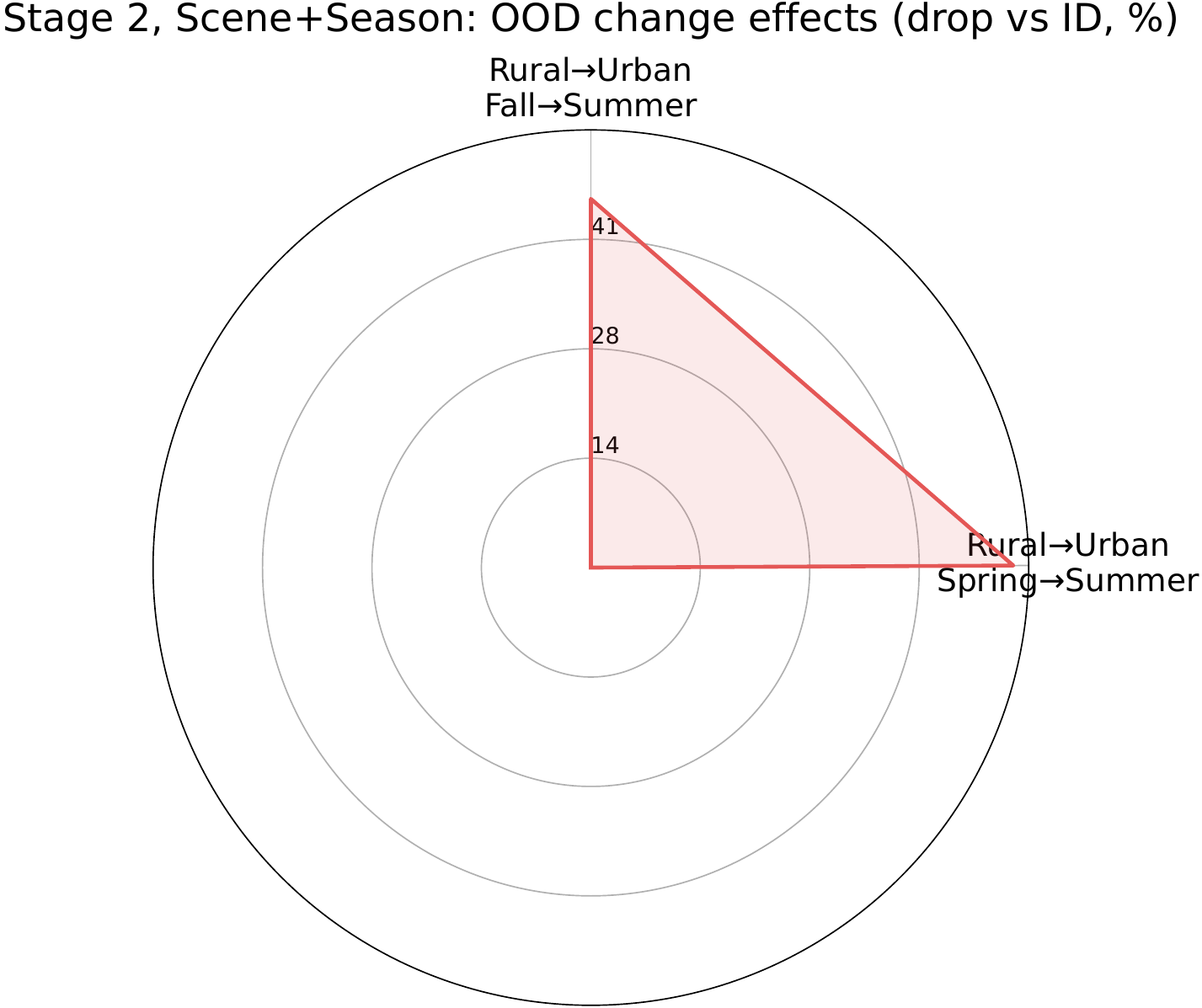}
  \caption{Stage 2, Scene, Season}
\end{subfigure}\hfill

\vspace{0.7ex}

\begin{subfigure}[t]{0.32\textwidth}
  \centering
  \includegraphics[height=0.19\textheight,keepaspectratio]{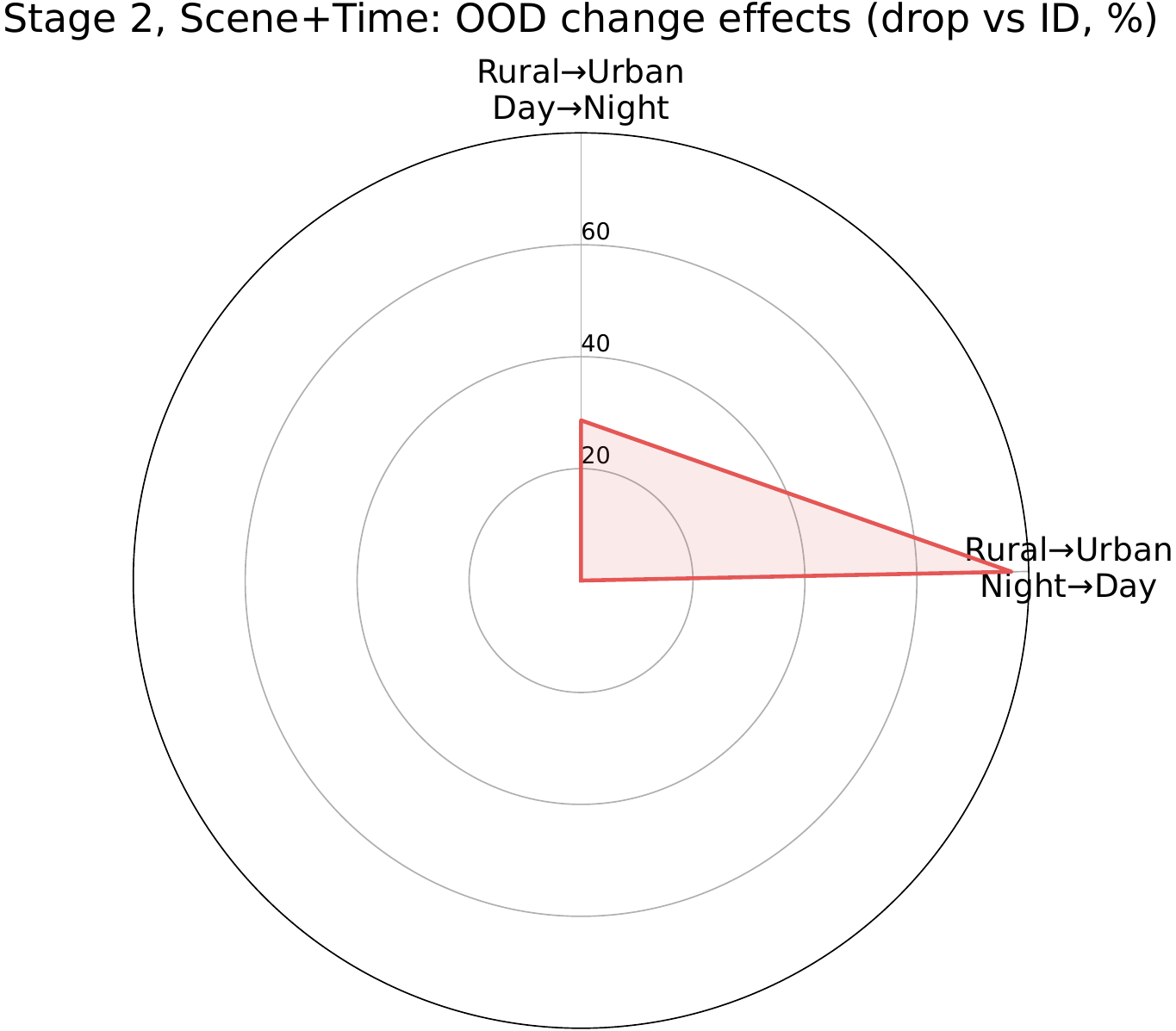}
  \caption{Stage 2, Scene, Time}
\end{subfigure}
\begin{subfigure}[t]{0.32\textwidth}
  \centering
  \includegraphics[height=0.19\textheight,keepaspectratio]{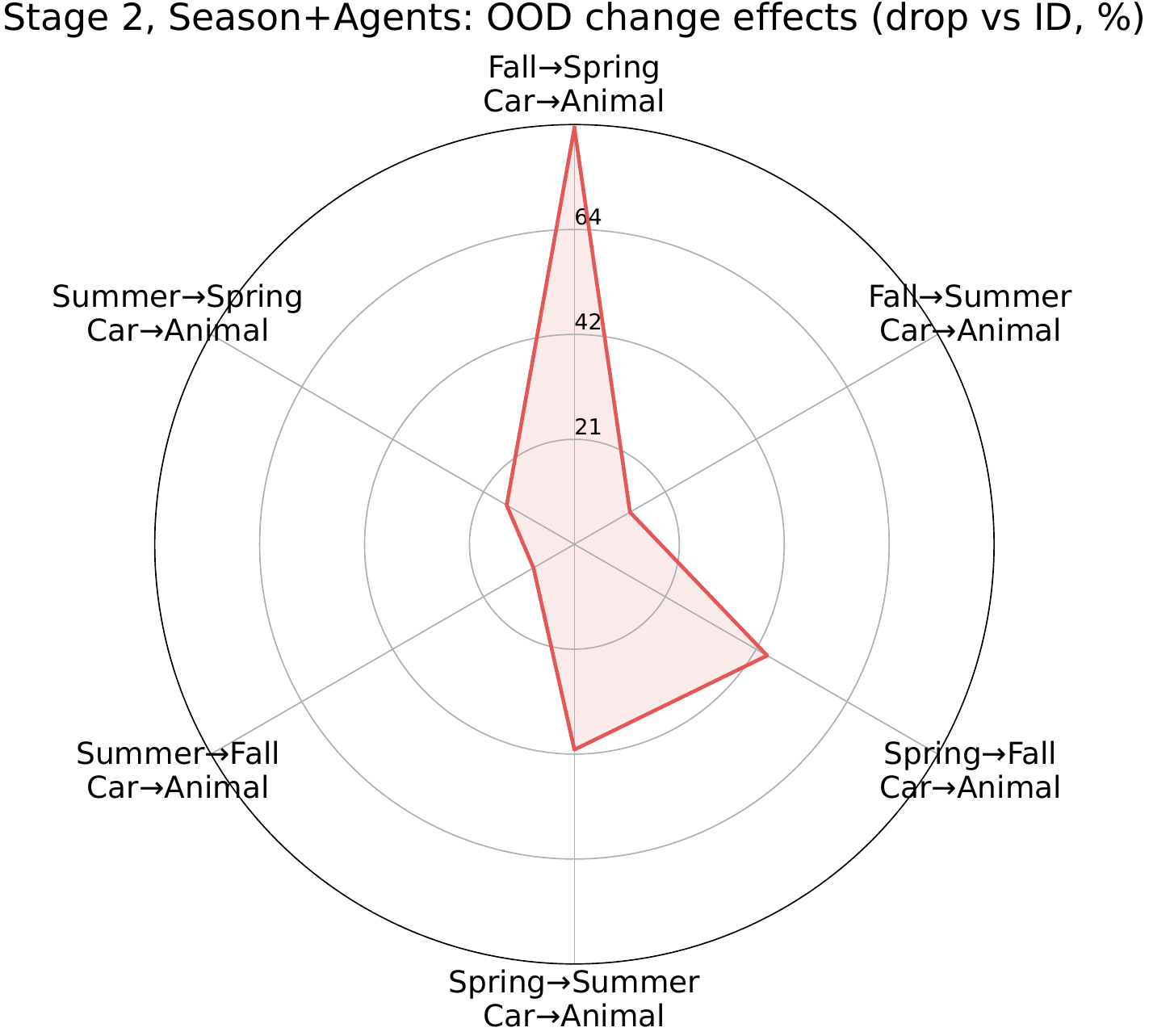}
  \caption{Stage 2, Season, Agents}
\end{subfigure}\hfill
\begin{subfigure}[t]{0.32\textwidth}
  \centering
  \includegraphics[height=0.19\textheight,keepaspectratio]{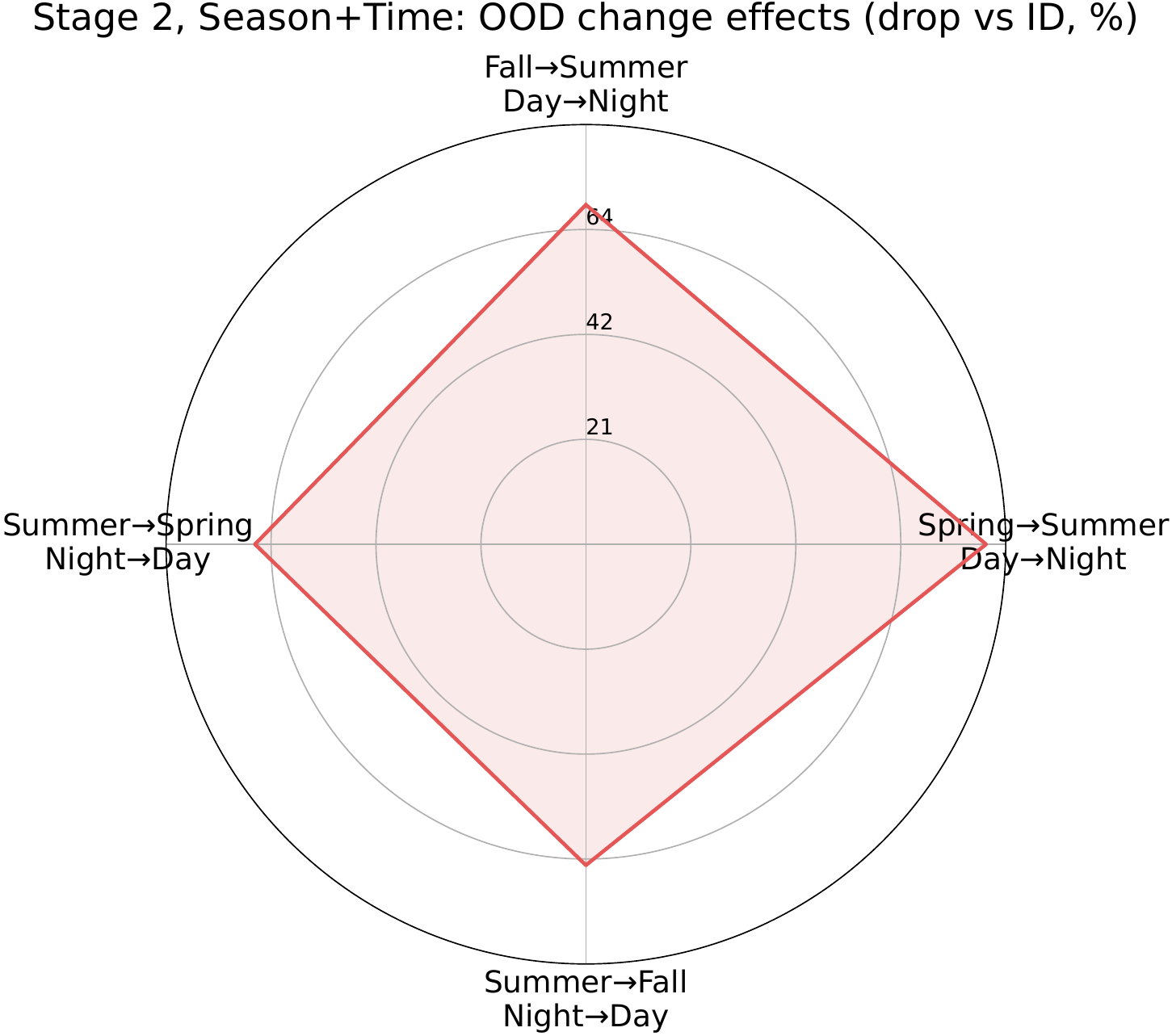}
  \caption{Stage 2, Season, Time}
\end{subfigure}\hfill

\vspace{0.7ex}

\begin{subfigure}[t]{0.32\textwidth}
  \centering
  \includegraphics[height=0.19\textheight,keepaspectratio]{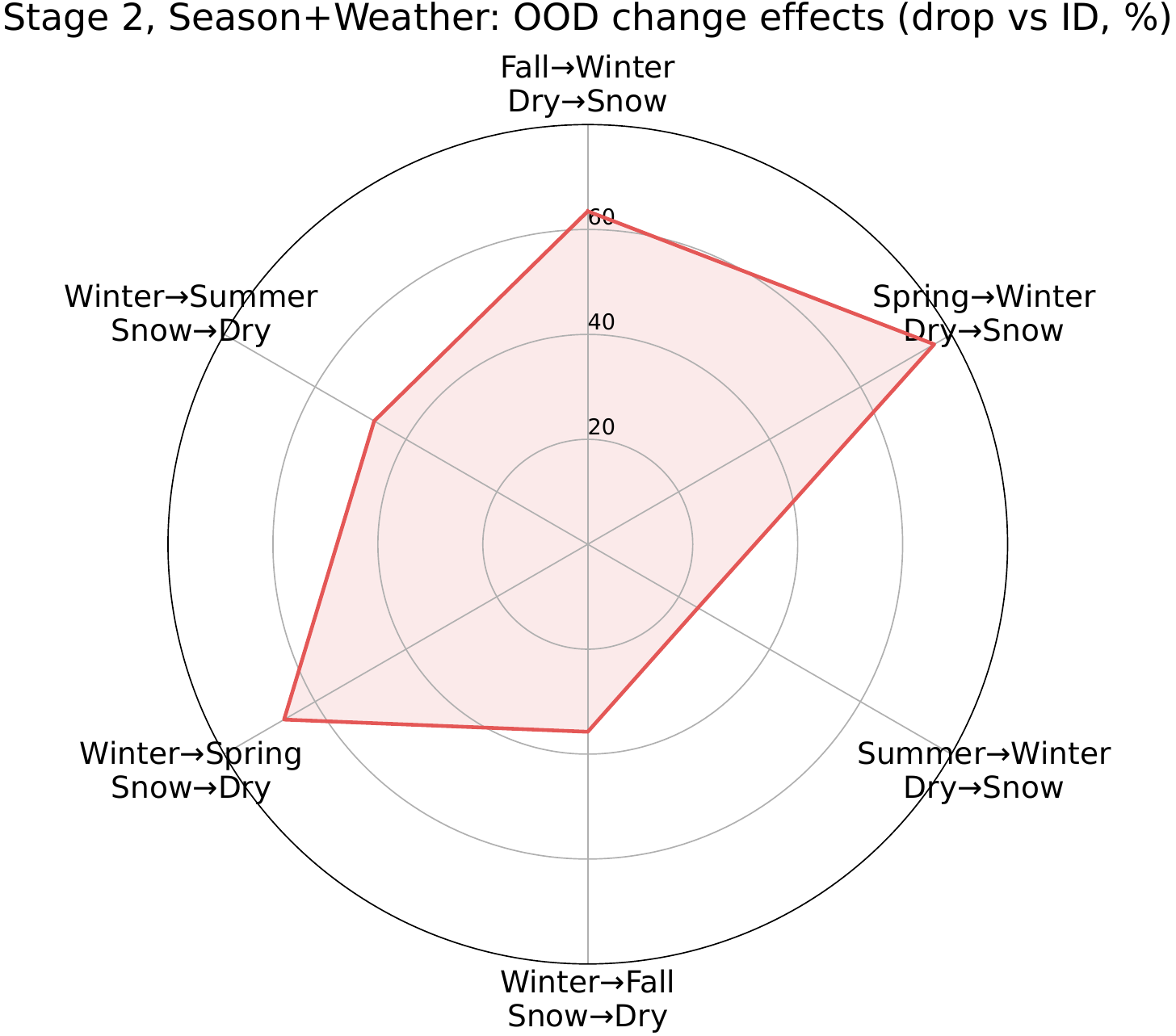}
  \caption{Stage 2, Season, Weather}
\end{subfigure}\hfill
\begin{subfigure}[t]{0.32\textwidth}
  \centering
  \includegraphics[height=0.19\textheight,keepaspectratio]{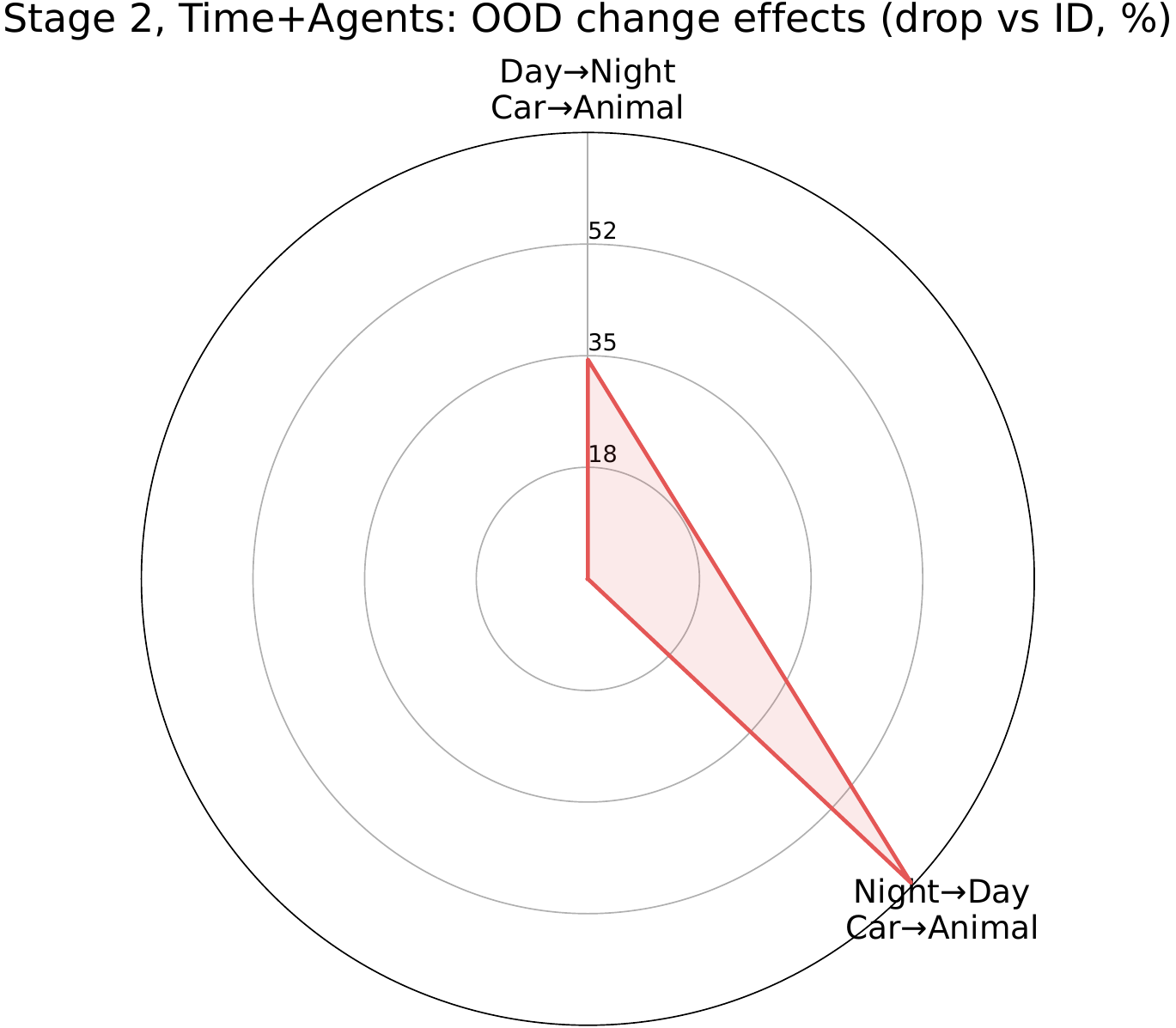}
  \caption{Stage 2, Time, Agents}
\end{subfigure}
\begin{subfigure}[t]{0.32\textwidth}
  \centering
  \includegraphics[height=0.19\textheight,keepaspectratio]{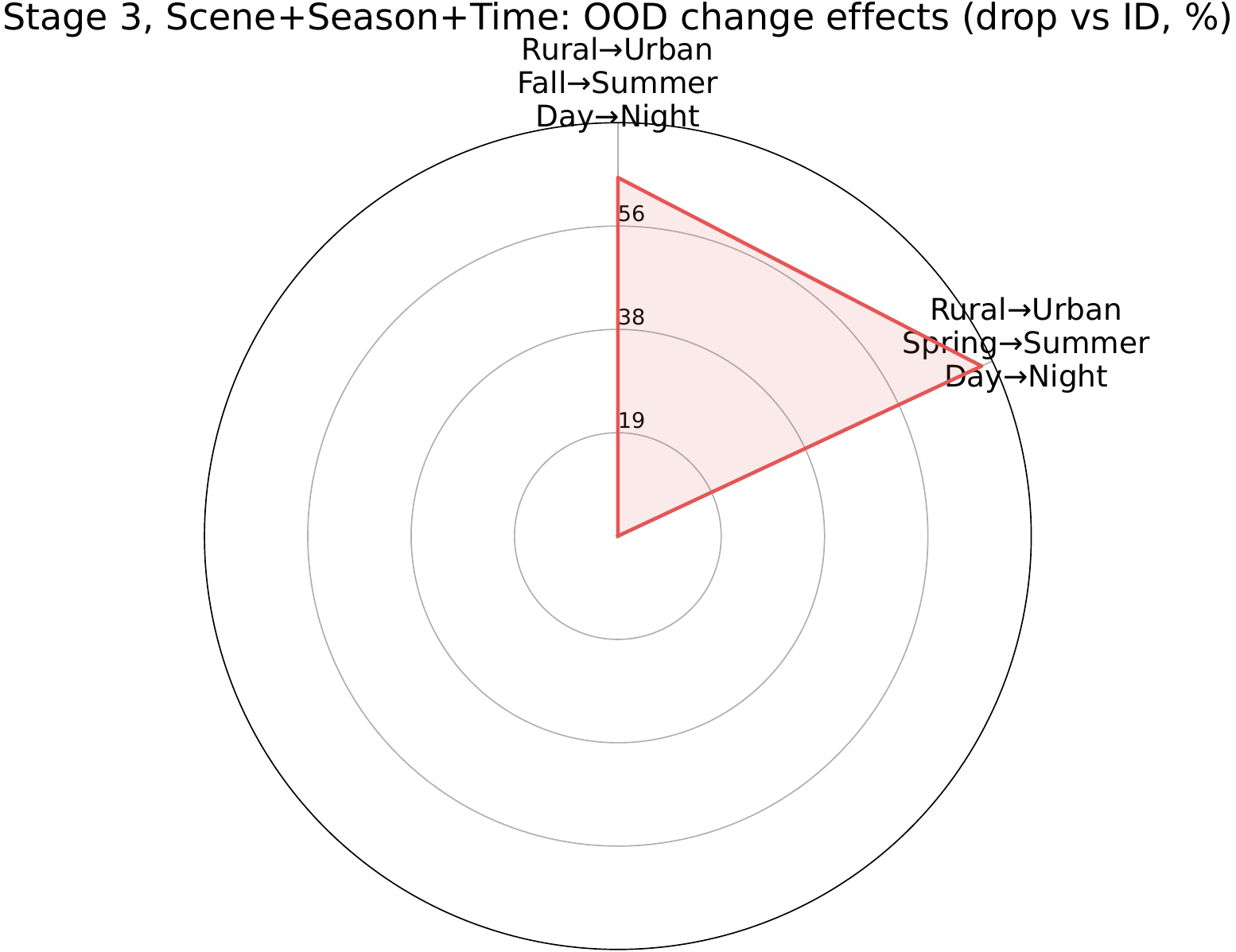}
  \caption{Stage 3, Scene, Season, Time}
\end{subfigure}\hfill

\vspace{0.7ex}

\begin{subfigure}[t]{0.32\textwidth}
  \centering
  \includegraphics[height=0.19\textheight,keepaspectratio]{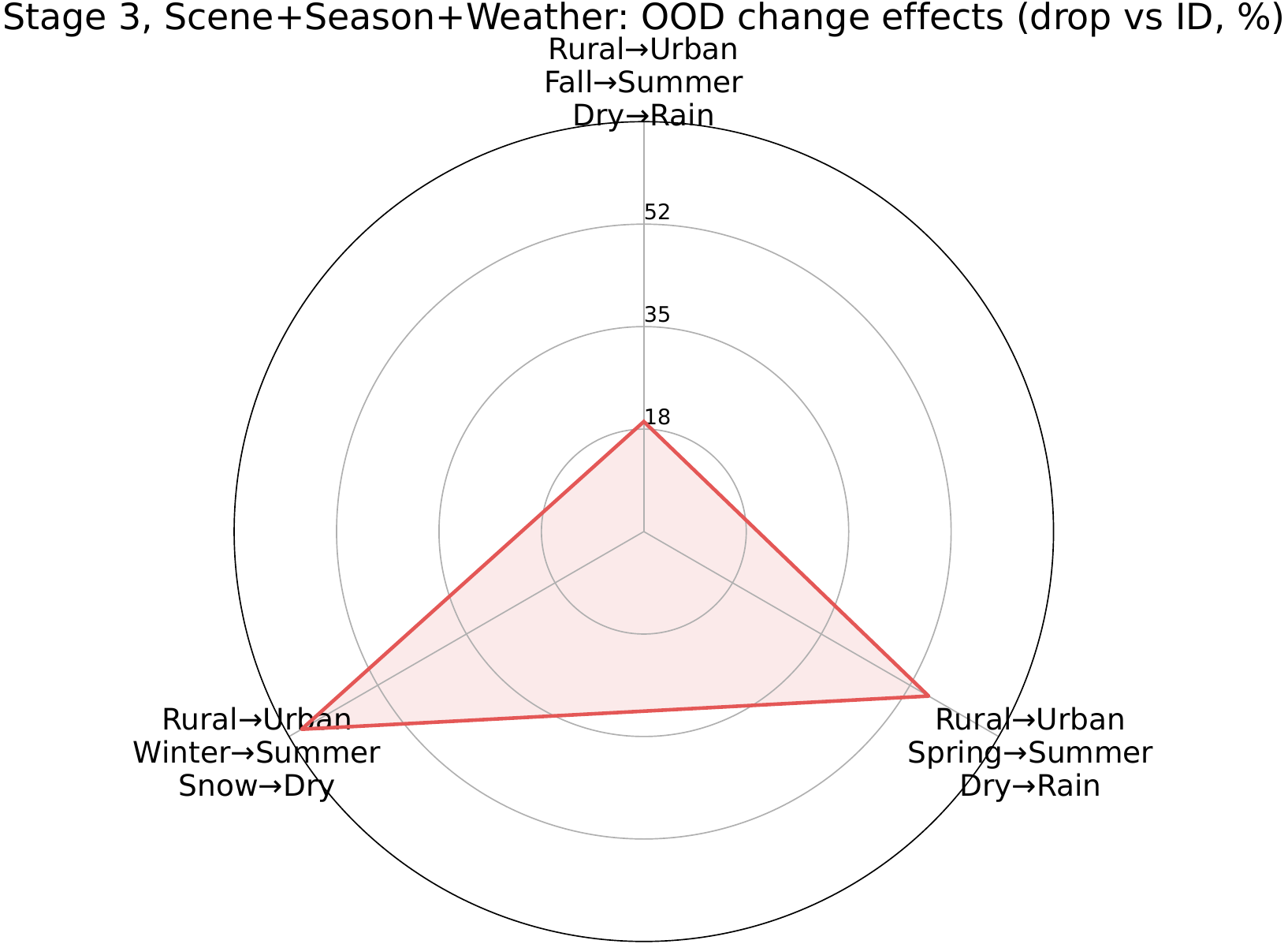}
  \caption{Stage 3, Scene, Season, Weather}
\end{subfigure}\hfill
\begin{subfigure}[t]{0.3\textwidth}
  \centering
  \includegraphics[height=0.19\textheight,keepaspectratio]{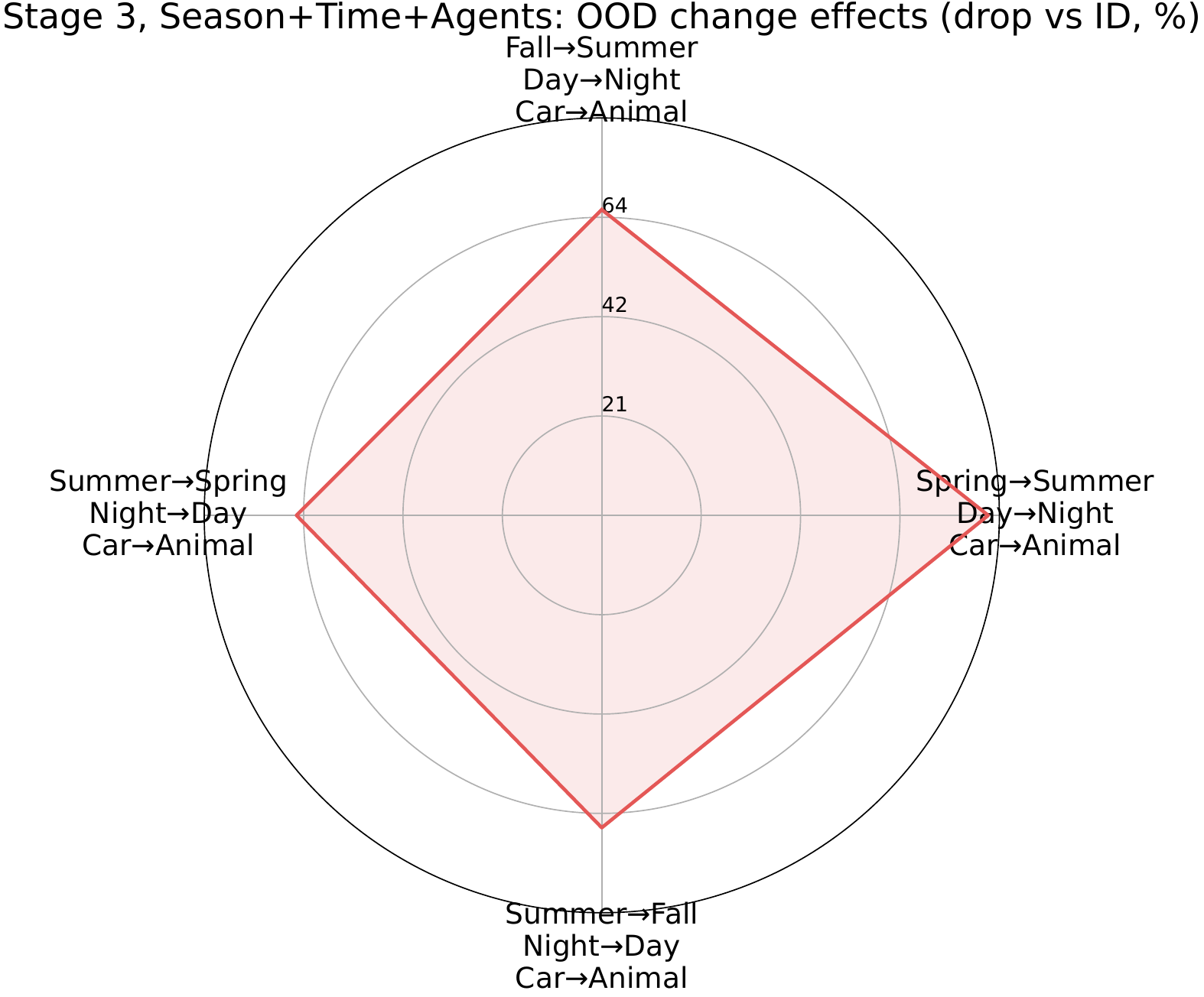}
  \caption{Stage 3, Season, Time, Agents}
\end{subfigure}\hfill
\begin{subfigure}[t]{0.3\textwidth}
  \centering
  \includegraphics[height=0.19\textheight,keepaspectratio]{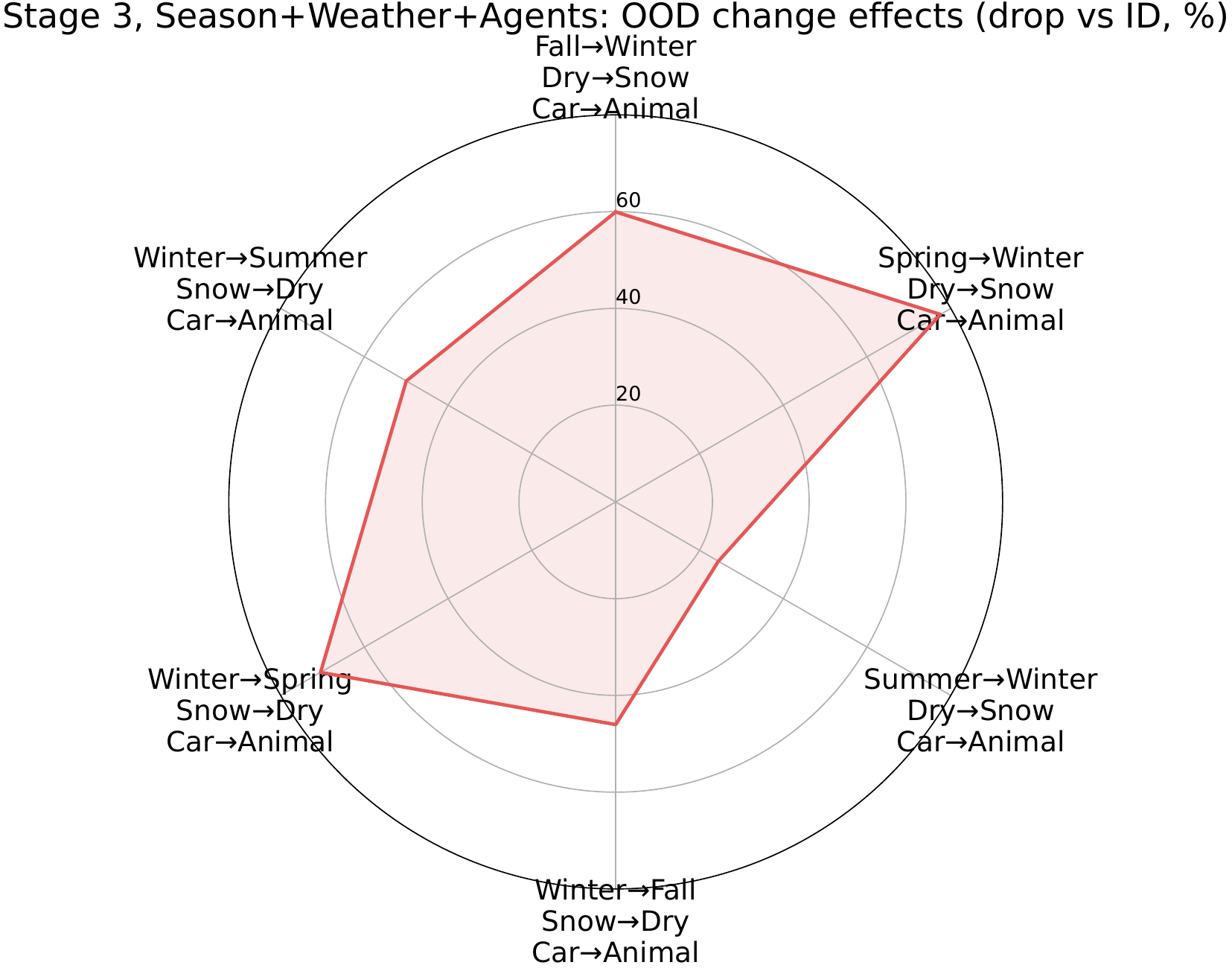}
  \caption{Stage 3, Season, Weather, Agents}
\end{subfigure}

\caption{Themed star plots for multi factor shifts. Each subplot aggregates all shifts matching the factor theme.}
\label{fig:themed_star_main}
\vspace{-1.0ex}
\end{figure*}

\textbf{Temporal context.}
Pass+ViT (MF, T=16) attains $49.2\%$, which is $-8.1$ points below Pass+ViT (single-frame, 14T,2V) at $57.2\%$.
Pass+RCNN (MF) reaches $47.0\%$, outperforming Pass+CNN ($40.4\%$) but still below both single-frame ViT variants ($57.2\%$ at 14T,2V; $52.4\%$ at 5T,1V). These results suggest:
\begin{tcolorbox}[colback=gray!10,colframe=black!50,title=Takeaway 3,
boxsep=2pt, left=3pt, right=3pt, top=5pt, bottom=5pt]
\textbf{Naïvely adding multi-frame context does not surpass the best single-frame baseline.}
\end{tcolorbox}
Thus, temporal aggregation must be designed carefully (e.g., alignment and motion-aware fusion) to translate into consistent OOD gains at comparable latency.

\begin{figure*}[t]
    \centering
\includegraphics[width=0.9\textwidth,height=0.35\textheight]{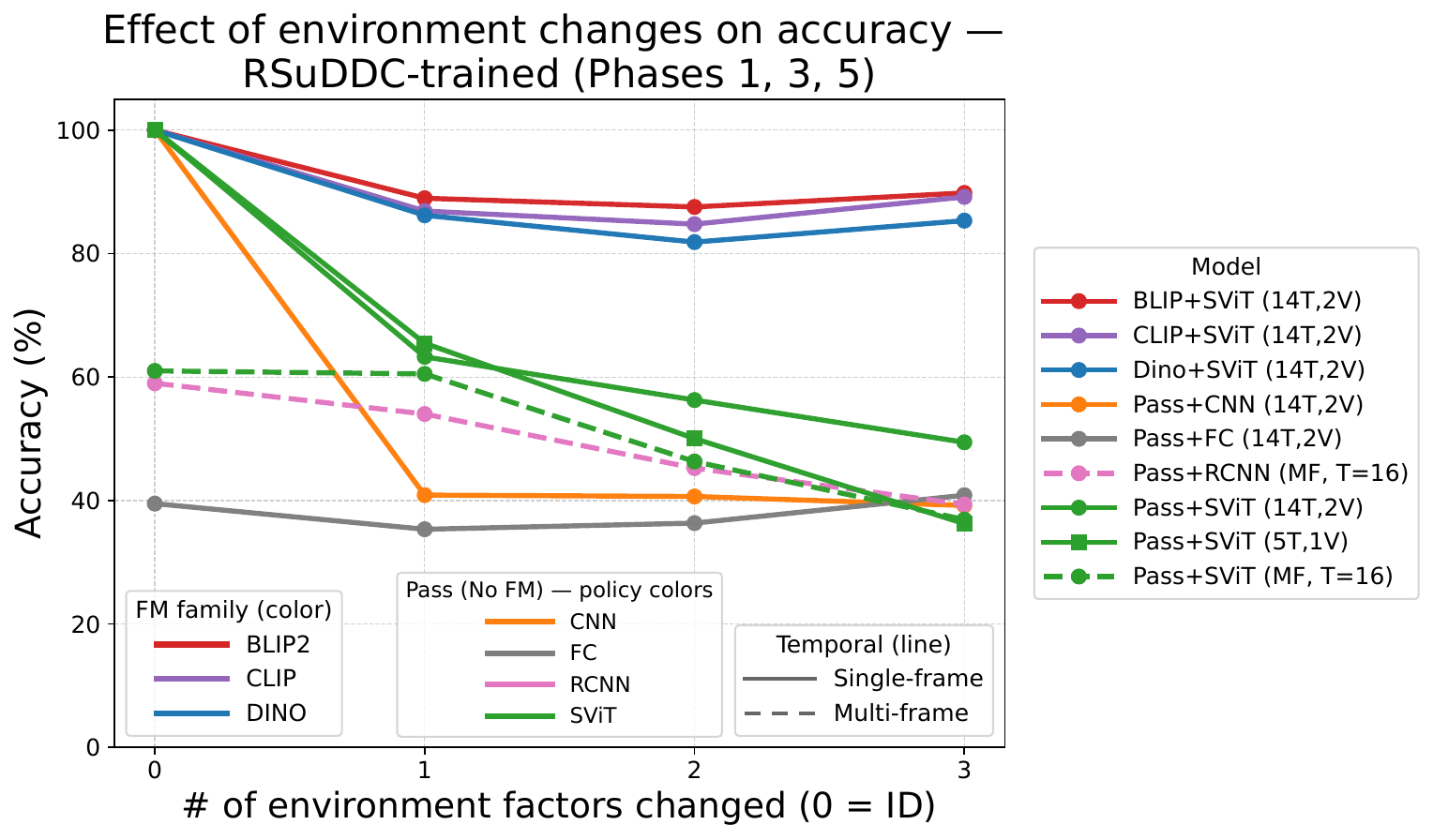}
\caption{Accuracy as a function of environment factors changes across model variants.}
\label{fig:arch_accuracy_vs_numchanges}
\end{figure*}

\subsection{OOD Environmental Factor Shifts and Their Effect}
\textbf{Which factors matter most?}
We quantify robustness by the \emph{drop} in closed-loop accuracy relative to each model’s in-distribution (ID) baseline. Single-factor shifts (Fig.~\ref{fig:EOC_Stage1}) reveal a clear ordering of difficulty: switching \textbf{scene} from rural$\rightarrow$urban and \textbf{time} from day$\rightarrow$night are the two dominant sources of degradation, each causing about a $31\%$ average drop (Scene: $31.15\%$, Time: $31.00\%$). 
In contrast, changing the \textbf{actor} from car$\rightarrow$animal is comparatively mild ($10.27\%$), and light \textbf{weather} variation (dry$\rightarrow$rain) is smallest ($6.86\%$).
Seasonal shifts are typically modest (e.g., summer$\rightarrow$fall: $9.62\%$, summer$\rightarrow$spring: $15.23\%$), but certain seasonal discontinuities are severe (fall$\rightarrow$spring: $84.63\%$). Large time reversals also hurt in the opposite direction (night$\rightarrow$day: $67.4\%$). These patterns are consistent with intuitive factors: urban scenes add visual density, clutter, and occlusion; illumination inversions (day$\leftrightarrow$night) fundamentally alter photometrics and signal-to-noise; by comparison, swapping the actor class perturbs semantics more than geometry and thus affects ego-motion control less; moderate rain impacts appearance but preserves most structural cues.

Two- (Fig.~\ref{fig:EOC_Stage2}) and three-factor (Fig.~\ref{fig:EOC_Stage3}) shifts amplify these trends. Pairings that \emph{include time flips} balloon the drop: season+time combinations (e.g., spring$\rightarrow$summer with day$\rightarrow$night) reach $81.0\%$, while fall$\rightarrow$summer with day$\rightarrow$night is $68.8\%$. Scene+time (rural$\rightarrow$urban with day$\rightarrow$night) remains challenging, but more moderate on average ($28.6\%$), reflecting partial complementarity between increased texture/occlusion and low-light noise. Triples that include a time inversion similarly dominate (e.g., rural$\rightarrow$urban + spring$\rightarrow$summer + day$\rightarrow$night: $72.9\%$; season+time+actor with spring$\rightarrow$summer + day$\rightarrow$night + car$\rightarrow$animal: $82.6\%$). In contrast, combinations \emph{without} time flips-such as scene+weather+actor (rural$\rightarrow$urban + dry$\rightarrow$rain + car$\rightarrow$animal: $22.3\%$) or season+weather+actor with summer$\rightarrow$winter and snow—remain substantially more manageable ($\approx\!24.5\%$). 

\begin{tcolorbox}[colback=gray!10,colframe=black!50,title=Takeaway 4,
boxsep=2pt, left=3pt, right=3pt, top=5pt, bottom=5pt]
The primary axes of brittleness are illumination (\emph{time}) and distributional density/geometry (\emph{scene}). Seasonal shifts matter when they induce large textural/illumination discontinuities (e.g., fall$\rightarrow$spring). Actor identity is least impactful for low-level control, and moderate precipitation alone is relatively benign.
\end{tcolorbox}

These results suggest: targeted data collection for nocturnal/low-light and dense urban conditions should yield the greatest robustness gains across architectures.
For a complete breakdown into themed star plots per stage, see Figure~\ref{fig:themed_star_main}, and Appendix~\ref{app:add_theme}, Figures~\ref{fig:themed_star_stage1_all} to~\ref{fig:themed_star_stage3_all}.

\textbf{How many changes can a policy tolerate?}
We summarize tolerance by averaging accuracy at each change level (0, 1, 2, 3 changed factors) across environments. As shown in Fig.~\ref{fig:arch_accuracy_vs_numchanges}, models with foundation-model (FM) features remain remarkably stable: BLIP+ViT holds at $88.96/87.55/89.82\%$ for 1/2/3 changes, CLIP+ViT at $86.89/84.77/89.18\%$, and DINO+ViT at $86.20/81.86/85.33\%$. In contrast, lightweight non-FM policies degrade steadily as changes accumulate: Pass+ViT (single-frame) drops from $100\%\!\to\!63.28\%\!\to\!56.28\%\!\to\!49.43\%$, Pass+CNN from $100\%\!\to\!40.88\%\!\to\!40.66\%\!\to\!39.20\%$, and the multi-frame variants converge near the high-$30$s by three changes (Pass+ViT MF: $61.0/60.51/46.32/36.91\%$; Pass+RCNN MF: $59.0/54.02/45.25/39.36\%$). Three patterns emerge: 

\begin{tcolorbox}[colback=gray!10,colframe=black!50,title=Takeaway 5,
boxsep=2pt, left=3pt, right=3pt, top=5pt, bottom=5pt]
(i) The \emph{first} factor change produces the dominant drop for non-FM single-frame models (e.g., Pass+ViT loses $36.7$ points at one change, with only modest additional declines thereafter); (ii) temporal (multi-frame) policies are more resilient to the first change, but their robustness erodes once a second change is introduced; and (iii) FM-based extractors maintain performance above $\sim\!85\%$ under two or three simultaneous shifts, whereas non-FM models fall below $50\%$ by the third change.
\end{tcolorbox}

\noindent\textbf{Are factor interactions additive?}
No, interactions are decidedly non-additive. As can be seen in Fig.~\ref{fig:arch_accuracy_vs_numchanges}, on FM models, accuracy at three changes occasionally \emph{rebounds} above the two-change level (e.g., BLIP+ViT: $87.55\%\!\to\!89.82\%$, CLIP+ViT: $84.77\%\!\to\!89.18\%$, DINO+ViT: $81.86\%\!\to\!85.33\%$), indicating that certain factor combinations are not simply “harder” in aggregate. Aggregated per-factor analyses reinforce this: a scene+time shift (rural$\!\to\!$urban with day$\!\to\!$night) yields a mean drop of $28.63\%$, which is \emph{less} than either scene alone ($31.15\%$) or time alone ($31.00\%$), suggesting partial compensation (e.g., urban lighting at night restoring salient edges). Conversely, pairings that involve \emph{season} with a time flip can be strongly super-additive (e.g., spring$\!\to\!$summer with day$\!\to\!$night: $81.02\%$ drop), and triples that stack season+time+\{actor/scene\} often remain severe (e.g., spring$\!\to\!$summer + day$\!\to\!$night + car$\!\to\!$animal: $82.62\%$). Taken together, 
\begin{tcolorbox}[colback=gray!10,colframe=black!50,title=Takeaway 6,
boxsep=2pt, left=3pt, right=3pt, top=5pt, bottom=5pt]
The first change imposes the largest penalty, and subsequent changes can either \emph{dampen} or \emph{amplify} difficulty depending on the axes involved—illumination flips (day$\leftrightarrow$night) are the key amplifier, while some scene+time pairings are partially mitigating.
\end{tcolorbox}

\subsection{Training Data choices}

\begin{figure}[!t]
  \centering

  \subfloat[Training ID vs accuracy\label{fig:training_id_vs_accuracy}]{
    \adjustbox{valign=t}{
      \includegraphics[width=\columnwidth,height=0.27\textheight,keepaspectratio]{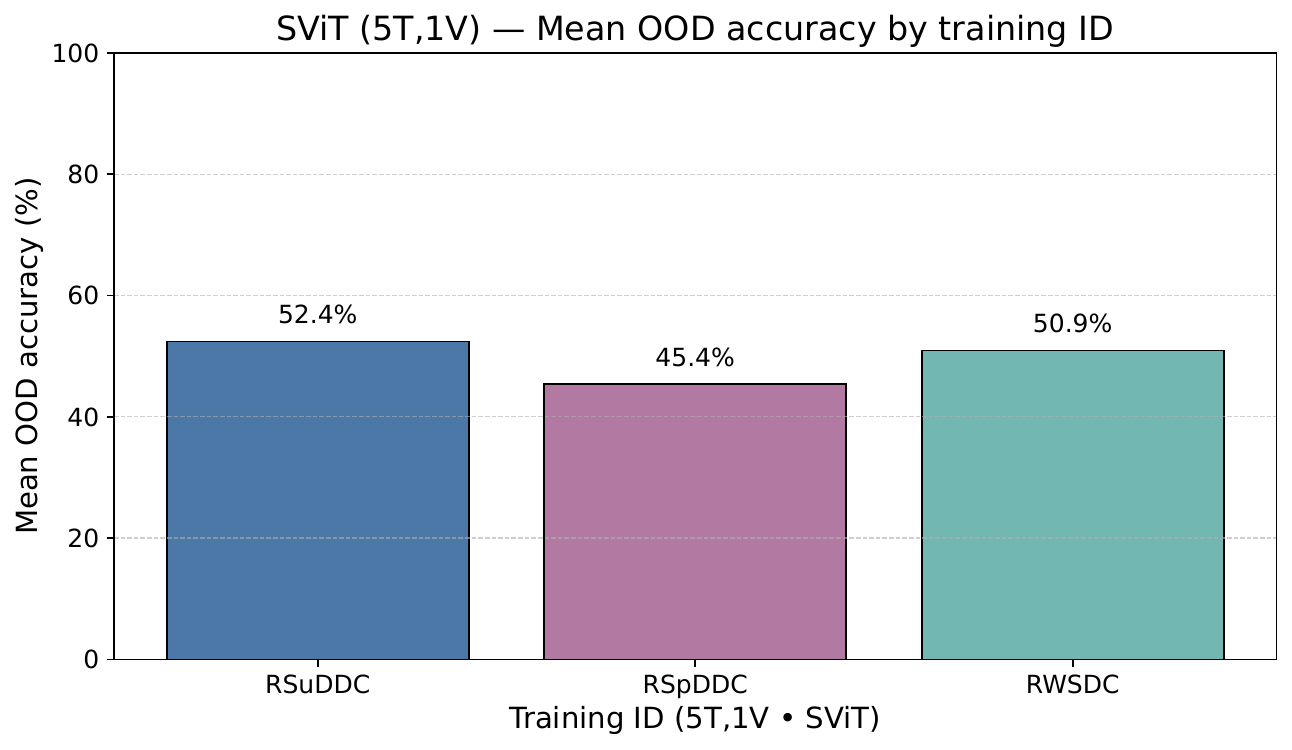}
    }
  }

  \vspace{0.35em}

  \subfloat[Environment changes vs accuracy\label{fig:env_stages_vs_accuracy}]{
    \adjustbox{valign=t}{
      \includegraphics[width=\columnwidth,height=0.27\textheight,keepaspectratio]{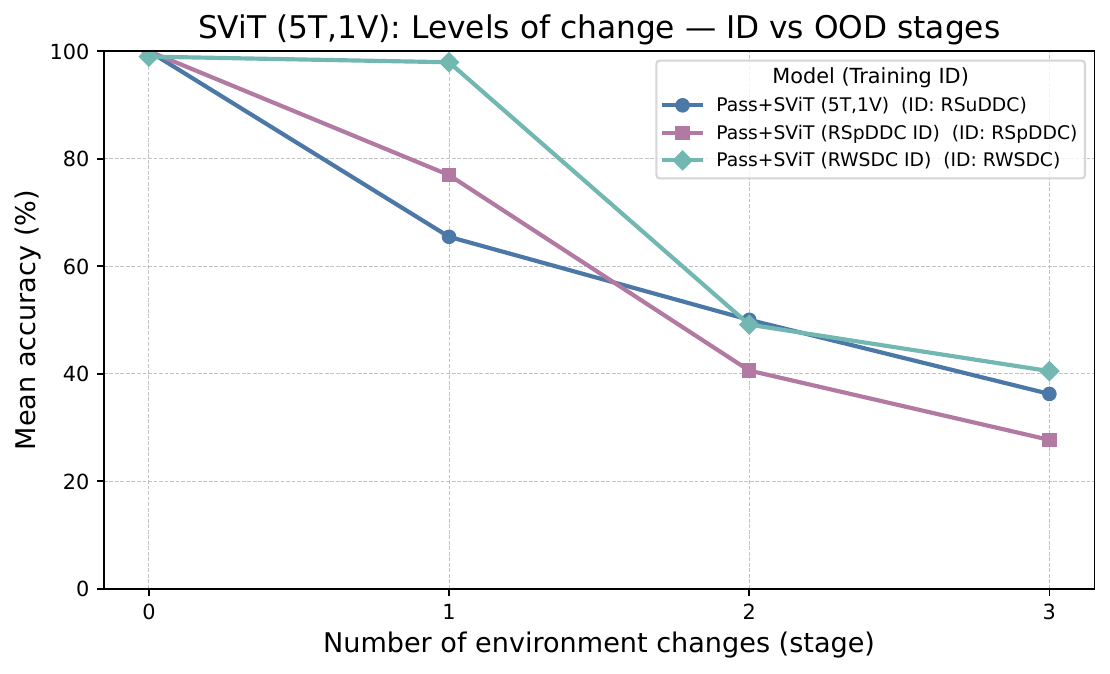}
    }
  }

  \vspace{0.35em}

  \subfloat[Training traces vs accuracy\label{fig:num_traces_vs_accuracy}]{
    \adjustbox{valign=t}{
      \includegraphics[width=\columnwidth,height=0.27\textheight,keepaspectratio]{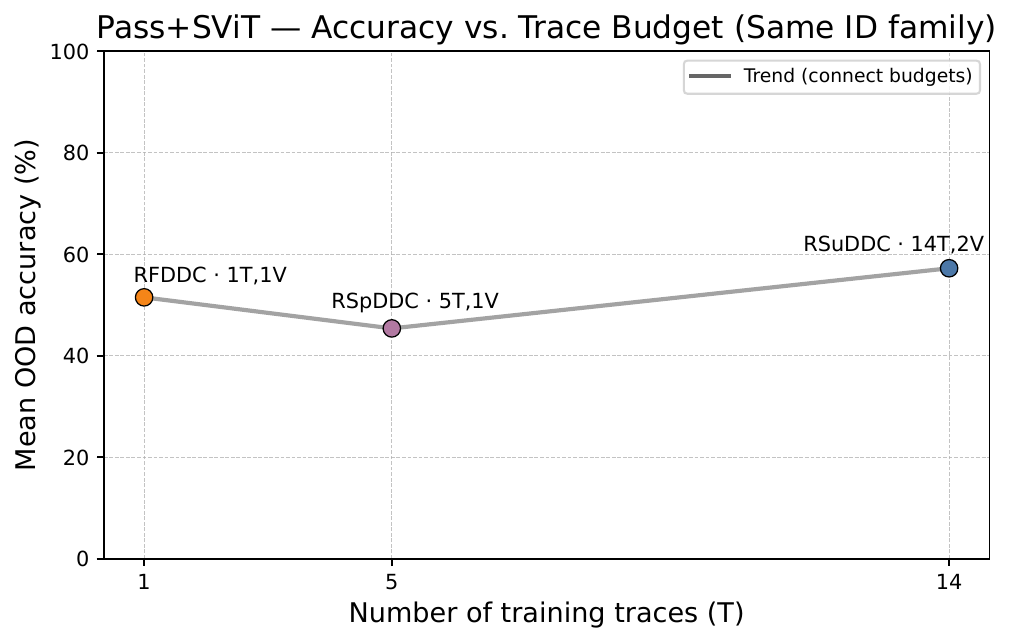}
    }
  }

  \caption{Accuracy trends under training distribution, environmental changes, and number of training traces.}
  \label{fig:task4_task5_task6}
\end{figure}

\textbf{Under which settings is it better to train a model?}
Controlling for architecture and budget \mbox{(ViT, 5T/1V)}, we compare three training IDs: \textit{RSuDDC} (rural–summer–dry–day–car), \textit{RSpDDC} (spring–dry–day), and \textit{RWSDC} (winter–snow–day). Averaged over all OODs (Fig.~\ref{fig:training_id_vs_accuracy}), \textbf{RSuDDC} yields the best overall robustness ($52.4\%$), followed closely by \textbf{RWSDC} ($50.9\%$) and then \textbf{RSpDDC} ($45.4\%$). Stage-wise trends (Fig.~\ref{fig:env_stages_vs_accuracy}) reveal complementary strengths. With \emph{one} factor changed, the model trained on \textbf{RWSDC} is strikingly resilient ($97.95\%$), outperforming both \textbf{RSpDDC} ($76.96\%$) and \textbf{RSuDDC} ($65.48\%$). This indicates that exposure to adverse winter/snow conditions promotes invariances that translate exceptionally well to single-axis shifts. As changes \emph{compound}, \textbf{RSuDDC} and \textbf{RWSDC} remain comparatively stable, while \textbf{RSpDDC} degrades fastest: at two changes the means are $50.04\%$ (RSuDDC), $49.14\%$ (RWSDC), vs.\ $40.60\%$ (RSpDDC); at three changes $36.25\%$ (RSuDDC), $40.49\%$ (RWSDC), vs.\ $27.70\%$ (RSpDDC). 

\begin{tcolorbox}[colback=gray!10,colframe=black!50,title=Takeaway 7,
boxsep=2pt, left=3pt, right=3pt, top=5pt, bottom=5pt]
\noindent\textbf{Takeaway.} If deployments are expected to see \emph{isolated} shifts (one factor at a time), training on winter/snow (\textbf{RWSDC}) provides the strongest single-change robustness with only a small tradeoff in global average. For \emph{compounded} shifts, both \textbf{RSuDDC} and \textbf{RWSDC} are safer choices than \textbf{RSpDDC}. For broad, mixed-condition operation where overall average matters most, \textbf{RSuDDC} remains the most reliable single-ID training choice.
\end{tcolorbox}

\textbf{Training data scale.}
Holding the architecture fixed (ViT) while varying the number of training traces (Fig.~\ref{fig:num_traces_vs_accuracy}) shows a clear—though not strictly monotonic—benefit from scale. Moving from \mbox{5T/1V} (\textbf{RSpDDC}) to \mbox{14T/2V} (\textbf{RSuDDC}) increases the mean OOD accuracy from $45.4\%$ to $57.2\%$ (\,$\uparrow\!11.8$ points). Even a \mbox{1T/1V} model (\textbf{RFDDC}) reaches $51.5\%$, outperforming the \mbox{5T/1V} \textbf{RSpDDC} model despite far fewer traces. This indicates that \emph{what} is seen can rival \emph{how much} is seen: scale helps, but content alignment (e.g., exposure to harder seasonal/appearance factors) can offset smaller budgets. Overall, when deployment budgets allow, higher trace count and view diversity (\mbox{14T/2V}) yields the strongest average robustness.

\begin{tcolorbox}[colback=gray!10,colframe=black!50,title=Takeaway 8,
boxsep=2pt, left=3pt, right=3pt, top=5pt, bottom=5pt]
More training traces and views improve robustness on average, but targeted exposure to challenging factors can partially substitute for scale.
\end{tcolorbox}

\begin{table}[!t]
  \centering
  \setlength{\tabcolsep}{4pt}
  \caption{Blip+Svit trained on 1 vs.\ 2 vs.\ 3 IDs: closed-loop success on ID and OODs (\%).}
  \label{tab:blip_svit_multiid}
  \newcolumntype{C}{>{\centering\arraybackslash}X}
  \begin{tabularx}{\columnwidth}{l c C C C}
    \toprule
    \makecell[l]{Env} & \makecell[c]{Stage} &
    \makecell[c]{Blip+Svit\\(1 Id: Rsuddc)} &
    \makecell[c]{Blip+Svit\\(2 Ids: Rsuddc+\\+Rsudnc)} &
    \makecell[c]{Blip+Svit\\(3 Ids: Rsuddc+\\+Rsudnc+\\+Usuddc)} \\
    \midrule
    RSuDDC (ID) & 0 & 100.0 & 98.3 & 98.9 \\
    RFDDC       & 1 &  96.6 & 98.6 & 98.5 \\
    RSpDDC      & 1 &  98.2 & 97.1 & 96.2 \\
    RSuDDA      & 1 &  97.3 & 96.2 & 96.9 \\
    RSuDNA      & 1 & \NA   & 80.8 & 85.7 \\
    RSuDNC      & 1 &  92.1 & \NA  & \NA  \\
    USuDDA      & 1 & \NA   & \NA  & 87.5 \\
    USuDDC      & 1 &  60.6 & 70.1 & \NA  \\
    USuDNA      & 1 & \NA   & \NA  & 86.8 \\
    USuDNC      & 1 & \NA   & 89.2 & \NA  \\
    USuRDC      & 1 & \NA   & \NA  & 93.1 \\
    \midrule
    RFDDA       & 2 &  99.2 & 94.7 & 98.7 \\
    RSpDDA      & 2 &  99.3 & 96.5 & 96.7 \\
    RSuDNA      & 2 &  86.2 & \NA  & 85.7 \\
    RWSDC       & 2 &  99.1 & 99.4 & 94.1 \\
    USuDDA      & 2 &  63.6 & \NA  & \NA  \\
    USuDNC      & 2 &  84.8 & \NA  & \NA  \\
    USuRDA      & 2 & \NA   & \NA  & 94.1 \\
    USuRDC      & 2 &  80.7 & 78.7 & \NA  \\
    \midrule
    RWSDA       & 3 &  98.7 & 99.4 & 97.7 \\
    USuDNA      & 3 &  86.0 & \NA  & \NA  \\
    USuRDA      & 3 &  84.8 & 81.0 & \NA  \\
    \bottomrule
  \end{tabularx}
\end{table}

\textbf{Training data diversity.}
Diversity across IDs (BLIP+ViT; 1 vs.\ 2 vs.\ 3 IDs) - See Table~\ref{tab:blip_svit_multiid}, trades a small loss on the nominal ID for broader OOD gains. On the ID itself (RSuDDC), accuracy is $100\%$ (1 ID), $98.3\%$ (2 IDs), and $98.9\%$ (3 IDs). By contrast, several OODs improve with diversity: \mbox{RFDDC} rises from $96.6\%$ (1 ID) to $98.6\%$ (2 IDs), and \mbox{USuDDC} jumps from $60.6\%$ (1 ID) to $70.1\%$ (2 IDs). Diversity also unlocks strong performance on previously challenging domains—e.g., \mbox{USuRDC} appears at $93.1\%$ for the 3-ID model in stage 1. For comparison, in stage 2 the same OOD appears for the 1-ID and 2-ID models at $80.7\%$ and $78.7\%$, respectively, which is consistent with the trend that modest ID diversity can unlock stronger robustness on urban regional shifts. Some shifts see mild regressions (e.g., \mbox{RSpDDA}: $99.3\%\!\rightarrow\!96.5\%\!\rightarrow\!96.7\%$; \mbox{RWSDC}: $99.1\%\!\rightarrow\!99.4\%\!\rightarrow\!94.1\%$), consistent with finite-capacity trade-offs when fitting multiple distributions. Net effect: multi-ID training broadens coverage and raises performance on previously weak axes (urban/\;USu), while incurring small drops on a few well-covered seasonal/time variants.

\begin{tcolorbox}[colback=gray!10,colframe=black!50,title=Takeaway 9,
boxsep=2pt, left=3pt, right=3pt, top=5pt, bottom=5pt]
Multi-ID training broadens coverage and lifts weak axes (urban and USu), while only slightly reducing accuracy on the nominal ID.
\end{tcolorbox}

\textbf{Specialization vs.\ generalization.}
As can be seen from Table~\ref{tab:blip_svit_multiid}, single-ID training excels on closely related conditions (e.g., \mbox{RSpDDA} at $99.3\%$ with 1 ID), reflecting sharp specialization around the nominal domain. Multi-ID training, however, delivers stronger \emph{generalization} along axes underrepresented by the base ID—e.g., substantial gains on urban and USu variants (\mbox{USuDDC}: $60.6\%\!\rightarrow\!70.1\%$ with 2 IDs; new \mbox{USuRDC} at $93.1\%$ with 3 IDs). In practice, if deployment environments are concentrated near a single domain, specialization preserves peak ID performance. When coverage across diverse scene/region factors is paramount, modest multi-ID diversity yields broader robustness with only a slight reduction on the nominal ID.

\begin{figure*}[t]
    \centering
\includegraphics[width=1\textwidth,height=0.4\textheight]{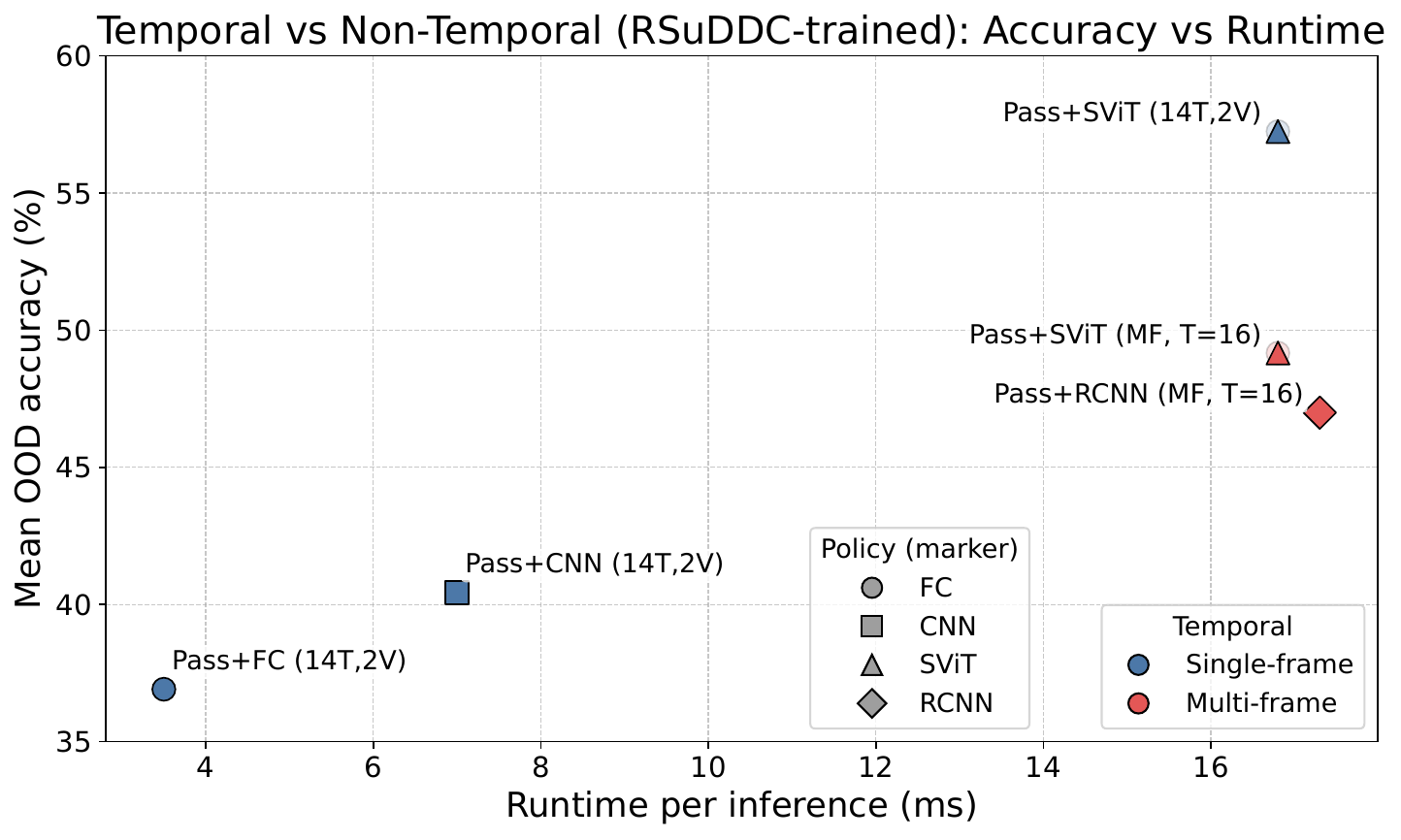}
\caption{Temporal vs Non-Temporal: Run-Time vs accuracy.}
\label{fig:temporal_vs_non_temporal}
\end{figure*}

\begin{tcolorbox}[colback=gray!10,colframe=black!50,title=Takeaway 10,
boxsep=2pt, left=3pt, right=3pt, top=5pt, bottom=5pt]
If deployment is concentrated near one domain, specialization preserves peak ID performance. If coverage across diverse scene and region factors matters, modest multi-ID diversity yields broader robustness with minimal ID loss.
\end{tcolorbox}

\subsection{Temporal Information}
Holding the Pass backbone fixed and comparing models at comparable runtime (See Fig.~\ref{fig:temporal_vs_non_temporal}), the single-frame ViT attains the highest mean OOD accuracy, \(57.24\%\) at \(16.8\,\mathrm{ms}\). The multi-frame ViT, at the same runtime, reaches \(49.17\%\) \(({-}8.07\) points relative to single-frame ViT\(),\) and the multi-frame RCNN at \(17.3\,\mathrm{ms}\) reaches \(46.99\%\) \(({-}10.25\) points\().\) Temporal models nevertheless improve substantially over the simpler single-frame baselines: MF-ViT exceeds FC \((36.91\%)\) and CNN \((40.44\%)\) by \(+12.26\) and \(+8.73\) points, respectively, while MF-RCNN exceeds them by \(+10.08\) and \(+6.55\) points. Overall, in this setting the strongest results come from a high-quality single-frame ViT, with temporal fusion yielding consistent gains over FC/CNN but not surpassing the single-frame ViT at matched compute.

\begin{tcolorbox}[colback=gray!10,colframe=black!50,title=Takeaway 11, 
boxsep=2pt, left=3pt, right=3pt, top=5pt, bottom=5pt]
At similar runtime, single-frame ViT delivers the best mean OOD accuracy \((57.24\%)\). Multi-frame models provide sizable gains over FC/CNN \((\approx\!+10\) points on average\(),\) but do not outperform the single-frame ViT in this training regime.
\end{tcolorbox}

\section{Conclusion}

We evaluate out-of-distribution robustness in vision-based driving by factorizing scene, season, weather, time, and agent mix, and measuring performance under controlled $k$-factor shifts. In VISTA closed-loop tests, ViT heads on BLIP-2, CLIP, or DINO features achieved top OOD accuracy (\(88.5\%\), \(86.4\%\), \(84.0\%\)), staying above \(85\%\) under three shifts, while non-FM baselines fell below \(50\%\). Not all factors are equal: rural$\to$urban and day$\to$night each caused $\sim\!31\%$ drops, actor swaps $\sim\!10\%$, and light rain $\sim\!7\%$, with seasonal flips sometimes severe. Interactions were non-additive: scene+time could be sub-additive, season+time often compounded. Temporal aggregation (Pass+ViT MF \(49.2\%\)) did not surpass the best single-frame baseline (\(57.2\%\)), though it beat FC/CNN. Training scale and design also mattered: more traces improved robustness (\(+11.8\) from 5 to 14), targeted hard cases substituted for scale, and multi-ID training broadened coverage (urban OOD \(60.6\%\!\to\!70.1\%\)) at minor ID cost. Overall, these results yield practical rules for building reliable closed-loop driving policies under multi-factor shifts.
Limitations include the use of simulation and a coarse discrete factorization, future work should validate the main takeaways in real world driving and refine the factors to finer, potentially continuous levels.

\bibliography{main}

\begin{thebibliography}{10}
\providecommand{\url}[1]{#1}
\csname url@samestyle\endcsname
\providecommand{\newblock}{\relax}
\providecommand{\bibinfo}[2]{#2}
\providecommand{\BIBentrySTDinterwordspacing}{\spaceskip=0pt\relax}
\providecommand{\BIBentryALTinterwordstretchfactor}{4}
\providecommand{\BIBentryALTinterwordspacing}{\spaceskip=\fontdimen2\font plus
\BIBentryALTinterwordstretchfactor\fontdimen3\font minus \fontdimen4\font\relax}
\providecommand{\BIBforeignlanguage}[2]{{%
\expandafter\ifx\csname l@#1\endcsname\relax
\typeout{** WARNING: IEEEtran.bst: No hyphenation pattern has been}%
\typeout{** loaded for the language `#1'. Using the pattern for}%
\typeout{** the default language instead.}%
\else
\language=\csname l@#1\endcsname
\fi
#2}}
\providecommand{\BIBdecl}{\relax}
\BIBdecl

\bibitem{Pomerleau1989}
D.~A. Pomerleau, ``Alvinn: An autonomous land vehicle in a neural network,'' in \emph{Advances in Neural Information Processing Systems (NeurIPS)}, vol.~1, 1989, pp. 305--313.

\bibitem{Bojarski2016}
M.~Bojarski, D.~D. Testa, D.~Dworakowski, B.~Firner, B.~Flepp, P.~Goyal, L.~D. Jackel, M.~Monfort, U.~Muller, J.~Zhang, X.~Zhang, J.~Zhao, and K.~Zieba, ``End to end learning for self-driving cars,'' \emph{arXiv:1604.07316}, 2016.

\bibitem{Chen2015DeepDriving}
C.~Chen, A.~Seff, A.~Kornhauser, and J.~Xiao, ``Deepdriving: Learning affordance for direct perception in autonomous driving,'' in \emph{IEEE International Conference on Computer Vision (ICCV)}, 2015, pp. 2722--2730.

\bibitem{amini2019variational}
A.~Amini, G.~Rosman, S.~Karaman, and D.~Rus, ``Variational end-to-end navigation and localization,'' in \emph{2019 International Conference on Robotics and Automation (ICRA)}.\hskip 1em plus 0.5em minus 0.4em\relax IEEE, 2019, pp. 8958--8964.

\bibitem{wang2023learning}
T.-H. Wang, W.~Xiao, M.~Chahine, A.~Amini, R.~Hasani, and D.~Rus, ``Learning stability attention in vision-based end-to-end driving policies,'' in \emph{Learning for Dynamics and Control Conference}.\hskip 1em plus 0.5em minus 0.4em\relax PMLR, 2023, pp. 1099--1111.

\bibitem{xiao2023barriernet}
W.~Xiao, T.-H. Wang, R.~Hasani, M.~Chahine, A.~Amini, X.~Li, and D.~Rus, ``Barriernet: Differentiable control barrier functions for learning of safe robot control,'' \emph{IEEE Transactions on Robotics}, 2023.

\bibitem{Codevilla2018}
F.~Codevilla, M.~M{\"u}ller, A.~L{\'o}pez, V.~Koltun, and A.~Dosovitskiy, ``End-to-end driving via conditional imitation learning,'' in \emph{IEEE International Conference on Robotics and Automation (ICRA)}, 2018, pp. 4693--4700.

\bibitem{Codevilla2019}
F.~Codevilla, E.~Santana, A.~M. L{\'o}pez, and A.~Gaidon, ``Exploring the limitations of behavior cloning for autonomous driving,'' in \emph{IEEE/CVF International Conference on Computer Vision (ICCV)}, 2019, pp. 9329--9338.

\bibitem{Tobin2017}
J.~Tobin, R.~Fong, A.~Ray, J.~Schneider, W.~Zaremba, and P.~Abbeel, ``Domain randomization for transferring deep neural networks from simulation to the real world,'' in \emph{IEEE/RSJ International Conference on Intelligent Robots and Systems (IROS)}, 2017, pp. 23--30.

\bibitem{Radford2021CLIP}
A.~Radford, J.~W. Kim, C.~Hallacy, A.~Ramesh, G.~Goh, S.~Agarwal, G.~Sastry, A.~Askell, P.~Mishkin, J.~Clark, G.~Krueger, and I.~Sutskever, ``Learning transferable visual models from natural language supervision,'' in \emph{International Conference on Machine Learning (ICML)}, 2021, pp. 8748--8763.

\bibitem{Caron2021DINO}
M.~Caron, H.~Touvron, I.~Misra, H.~J{\'e}gou, J.~Mairal, P.~Bojanowski, and A.~Joulin, ``Emerging properties in self-supervised vision transformers,'' in \emph{IEEE/CVF International Conference on Computer Vision (ICCV)}, 2021, pp. 9650--9660.

\bibitem{Li2023BLIP2}
J.~Li, D.~Li, S.~Savarese, and S.~C.~H. Hoi, ``{BLIP-2}: Bootstrapping language-image pre-training with frozen image encoders and large language models,'' in \emph{International Conference on Machine Learning (ICML)}, 2023.

\bibitem{sreeram2025probing}
S.~Sreeram, T.-H. Wang, A.~Maalouf, G.~Rosman, S.~Karaman, and D.~Rus, ``Probing multimodal llms as world models for driving,'' \emph{IEEE Robotics and Automation Letters}, 2025.

\bibitem{Zhang2022YouTube}
Q.~Zhang, Z.~Peng, and B.~Zhou, ``Learning to drive by watching {YouTube} videos: Action-conditioned contrastive policy pretraining,'' in \emph{European Conference on Computer Vision (ECCV)}, 2022, pp. 111--128.

\bibitem{Chen2022LAV}
D.~Chen and P.~Kr{\"a}henb{\"u}hl, ``Learning from all vehicles,'' in \emph{IEEE/CVF Conference on Computer Vision and Pattern Recognition (CVPR)}, 2022, pp. 17\,222--17\,231.

\bibitem{Wang2024Drive}
T.-H. Wang, A.~Maalouf, W.~Xiao, Y.~Ban, A.~Amini, G.~Rosman, S.~Karaman, and D.~Rus, ``Drive anywhere: Generalizable end-to-end autonomous driving with multi-modal foundation models,'' in \emph{2024 IEEE International Conference on Robotics and Automation (ICRA)}, 2024, pp. 6687--6694.

\bibitem{mallak2026see}
A.~Mallak, E.~Aasi, S.~Sreeram, T.-H. Wang, D.~Rus, and A.~Maalouf, ``See less, drive better: Generalizable end-to-end autonomous driving via foundation models stochastic patch selection,'' \emph{arXiv preprint arXiv:2601.10707}, 2026.

\bibitem{maalouf2023follow}
A.~Maalouf, N.~Jadhav, K.~M. Jatavallabhula, M.~Chahine, D.~M. Vogt, R.~J. Wood, A.~Torralba, and D.~Rus, ``Follow anything: Open-set detection, tracking, and following in real-time,'' \emph{arXiv preprint arXiv:2308.05737}, 2023.

\bibitem{chahine2024flexendtoendtextinstructedvisual}
\BIBentryALTinterwordspacing
M.~Chahine, A.~Quach, A.~Maalouf, T.-H. Wang, and D.~Rus, ``Flex: End-to-end text-instructed visual navigation with foundation models,'' 2024. [Online]. Available: \url{https://arxiv.org/abs/2410.13002}
\BIBentrySTDinterwordspacing

\bibitem{chahine2025decentralized}
M.~Chahine, W.~Yang, A.~Maalouf, J.~Siriska, N.~Jadhav, D.~Vogt, S.~Gil, R.~Wood, and D.~Rus, ``Decentralized vision-based autonomous aerial wildlife monitoring,'' \emph{arXiv preprint arXiv:2508.15038}, 2025.

\bibitem{Dosovitskiy2017CARLA}
A.~Dosovitskiy, G.~Ros, F.~Codevilla, A.~Lopez, and V.~Koltun, ``Carla: An open urban driving simulator,'' in \emph{Conference on robot learning}.\hskip 1em plus 0.5em minus 0.4em\relax PMLR, 2017, pp. 1--16.

\bibitem{Amini2022VISTA2}
A.~Amini, T.~Wang, I.~Gilitschenski, W.~Schwarting, Z.~Liu, S.~Han, S.~Karaman, and D.~Rus, ``{VISTA} 2.0: An open, data-driven simulator for multimodal sensing and policy learning for autonomous vehicles,'' in \emph{IEEE International Conference on Robotics and Automation (ICRA)}, 2022, pp. 2419--2426.

\bibitem{Xu2017FCNLSTM}
H.~Xu, Y.~Gao, F.~Yu, and T.~Darrell, ``End-to-end learning of driving models from large-scale video datasets,'' in \emph{IEEE Conference on Computer Vision and Pattern Recognition (CVPR)}, 2017, pp. 2174--2182.

\bibitem{Hu2022STP3}
S.~Hu, L.~Chen, P.~Wu, H.~Li, J.~Yan, and D.~Tao, ``{ST-P3}: End-to-end vision-based autonomous driving via spatial-temporal feature learning,'' in \emph{European Conference on Computer Vision (ECCV)}, 2022, pp. 533--549.

\bibitem{Shao2023ReasonNet}
H.~Shao, L.~Wang, R.~Chen, S.~L. Waslander, H.~Li, and Y.~Liu, ``Reasonnet: End-to-end driving with temporal and global reasoning,'' in \emph{IEEE/CVF Conference on Computer Vision and Pattern Recognition (CVPR)}, 2023, pp. 13\,723--13\,733.

\bibitem{Zhu2021SurveyAD}
Z.~Zhu and H.~Zhao, ``A survey of deep rl and il for autonomous driving policy learning,'' \emph{IEEE Transactions on Intelligent Transportation Systems}, vol.~23, no.~9, pp. 14\,043--14\,065, 2022.

\bibitem{Chen2023E2ESurvey}
L.~Chen, P.~Wu, K.~Chitta, B.~Jaeger, A.~Geiger, and H.~Li, ``End-to-end autonomous driving: Challenges and frontiers,'' \emph{arXiv:2306.16927}, 2023.

\bibitem{amini2022vista}
A.~Amini, T.-H. Wang, I.~Gilitschenski, W.~Schwarting, Z.~Liu, S.~Han, S.~Karaman, and D.~Rus, ``Vista 2.0: An open, data-driven simulator for multimodal sensing and policy learning for autonomous vehicles,'' in \emph{2022 International Conference on Robotics and Automation (ICRA)}.\hskip 1em plus 0.5em minus 0.4em\relax IEEE, 2022, pp. 2419--2426.

\bibitem{amir2021deep}
S.~Amir, Y.~Gandelsman, S.~Bagon, and T.~Dekel, ``Deep {ViT} features as dense visual descriptors,'' \emph{arXiv preprint arXiv:2112.05814}, 2021.

\bibitem{jatavallabhula2023conceptfusion}
K.~M. Jatavallabhula, A.~Kuwajerwala, Q.~Gu, M.~Omama, T.~Chen, S.~Li, G.~Iyer, S.~Saryazdi, N.~Keetha, A.~Tewari \emph{et~al.}, ``Conceptfusion: Open-set multimodal 3d mapping,'' \emph{arXiv preprint arXiv:2302.07241}, 2023.

\end{thebibliography}
\bibliographystyle{IEEEtran}

\clearpage

\twocolumn[{%
  \centering
  {\LARGE\bfseries Appendix\par}
  \vspace{1.2ex}
}]

\appendices

\section{Reproducibility details}
We provide the evaluation protocol and implementation details needed to reproduce training and closed loop evaluation across all studies. Unless an ablation explicitly modifies them, shared settings are kept fixed across models and environments.

\subsection{Evaluation Metrics and Protocol Details}
\label{app:metrics_details}
This subsection provides the closed loop evaluation protocol and aggregation rules used throughout the paper, enabling direct reproduction of all reported metrics.

\textbf{Closed loop rollouts.}
All evaluations are performed in closed loop in VISTA at 30 frames per second, using a fixed control horizon of 200 simulator steps per episode. For each configuration, we evaluate 100 episodes on the in distribution environment and 100 episodes on the out of distribution environment, using the same route lists and random seeds across all compared models within a study.

\textbf{Route completion and success.}
We report \emph{route completion} as the fraction of the horizon traversed before termination, computed as \(\texttt{steps\_traveled} / 200\) when the episode terminates early, and \(1.0\) otherwise. We also report closed loop success where relevant, defined as completing the full horizon without an early termination event.

\textbf{Infractions.}
For each episode, we log simulator infraction indicators including collision, out of lane, off road, and stability violations when available. We aggregate these into per configuration \emph{infraction counts} by summing the corresponding episode level indicators across the 100 episodes, and we report mean and standard deviation across seeds and routes.

\textbf{Statistical testing.}
When comparing two models under the same routes and seeds, we perform paired statistical tests on matched episodes and apply Holm correction across the set of hypotheses in the comparison. Unless an ablation explicitly changes the training support, splits and evaluation routes are held fixed within each study.

\subsection{Implementation Details and Hyperparameters}
\label{app:impl_details}
This subsection summarizes the shared training and evaluation setup, including optimization, temporal clip construction, and compute environment, to support reproducibility across all studies.

\textbf{Framework and configuration.}
All runs use a unified PyTorch Lightning and Hydra codebase. Each experiment is fully specified by a base YAML configuration plus explicit command line overrides, ensuring that extractor, policy, temporal settings, and environment factor selections are reproducible from logged configs.

\textbf{Optimization.}
We train with AdamW and cosine learning rate decay, using a mean squared error objective on curvature prediction and early stopping based on validation MSE. Across studies, learning rates are fixed per study and shared across all compared models, with typical values of $1 \times 10^{-3}$ and $1 \times 10^{-4}$ depending on the training setting. Validation is performed at fixed step intervals, and checkpoints are saved using the same schedule for all models in the study.

\textbf{Temporal settings.}
For multi frame models, both training and evaluation operate on fixed length clips sampled at 30 frames per second. We report two temporal windows: a short window with \(T{=}9\) and stride 2, and a long window with \(T{=}16\) and stride 2. Clip construction is identical at train time and test time via the dataset and closed loop evaluator clip settings.

\textbf{Execution and hardware.}
We run on a university managed cluster with four NVIDIA A100 GPUs with 40GB memory each. Training time is approximately four days per full run. We use distributed execution when needed for simulator throughput, and set the PyTorch multiprocessing sharing strategy to \texttt{file\_system} to avoid shared memory limitations. Closed loop evaluation uses headless rendering on GPU when available.

\section{Additional themed factor shift figures}
\label{app:add_theme}
\addcontentsline{toc}{section}{Additional themed factor shift figures} 
We provide the full set of themed star plots for each stage, grouped by factor theme. Figure~\ref{fig:themed_star_stage1_all} shows single factor themes, Figure~\ref{fig:themed_star_stage2_all} shows double factor themes, and Figure~\ref{fig:themed_star_stage3_all} shows triple factor themes.

\subsection{Single factor themed star plots}
\begin{figure*}[!t]
\centering

\begin{subfigure}[t]{0.32\textwidth}
  \centering
  \includegraphics[height=0.24\textheight,keepaspectratio]{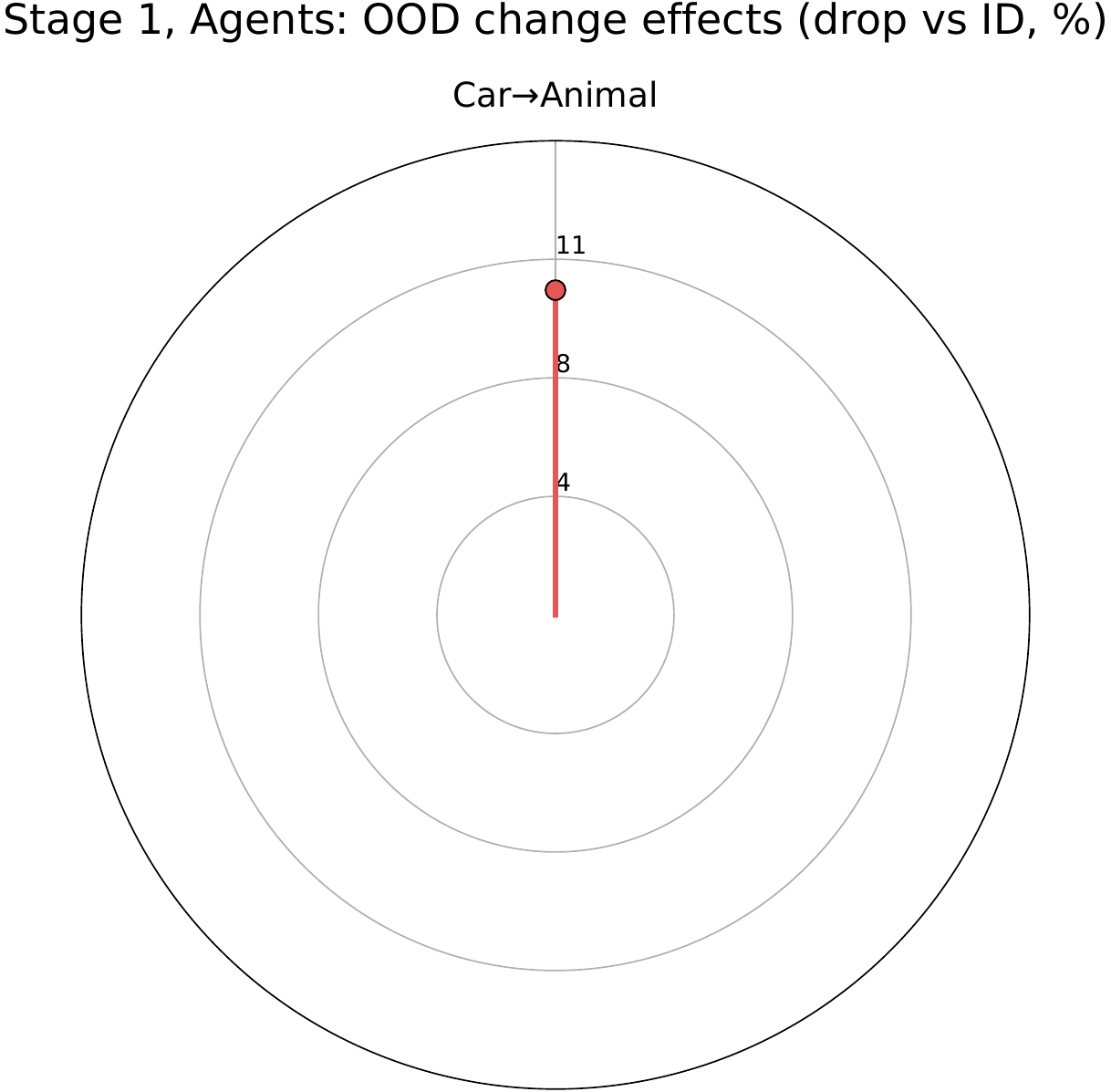}
  \caption{Agents}
\end{subfigure}\hfill
\begin{subfigure}[t]{0.32\textwidth}
  \centering
  \includegraphics[height=0.24\textheight,keepaspectratio]{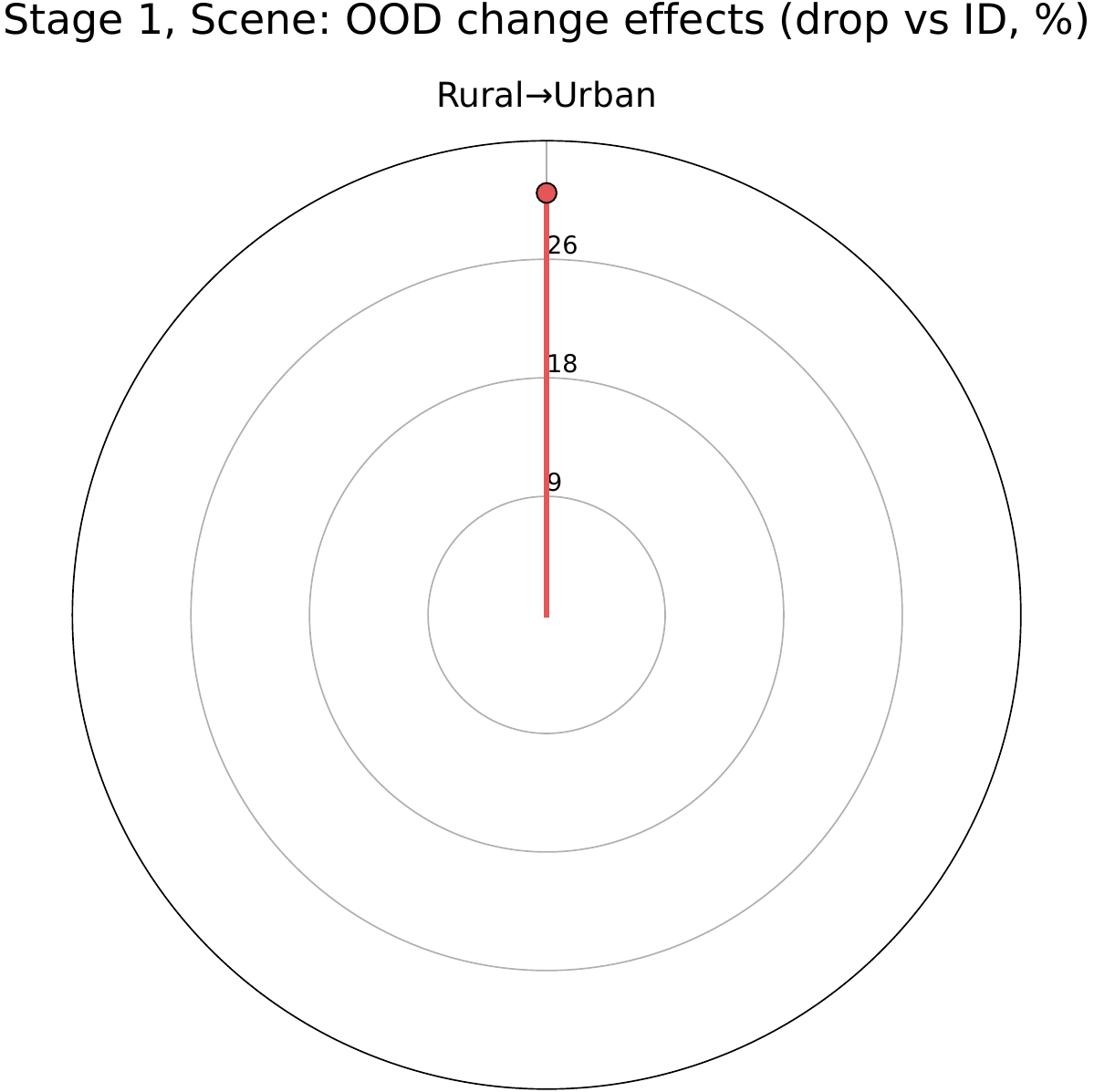}
  \caption{Scene}
\end{subfigure}\hfill
\begin{subfigure}[t]{0.32\textwidth}
  \centering
  \includegraphics[height=0.24\textheight,keepaspectratio]{Figures/Task3/Themes/Stage1/aggregate_stage1_theme_Season.pdf}
  \caption{Season}
\end{subfigure}

\vspace{0.8ex}

\begin{subfigure}[t]{0.32\textwidth}
  \centering
  \includegraphics[height=0.24\textheight,keepaspectratio]{Figures/Task3/Themes/Stage1/aggregate_stage1_theme_Time.pdf}
  \caption{Time}
\end{subfigure}\hfill
\begin{subfigure}[t]{0.32\textwidth}
  \centering
  \includegraphics[height=0.24\textheight,keepaspectratio]{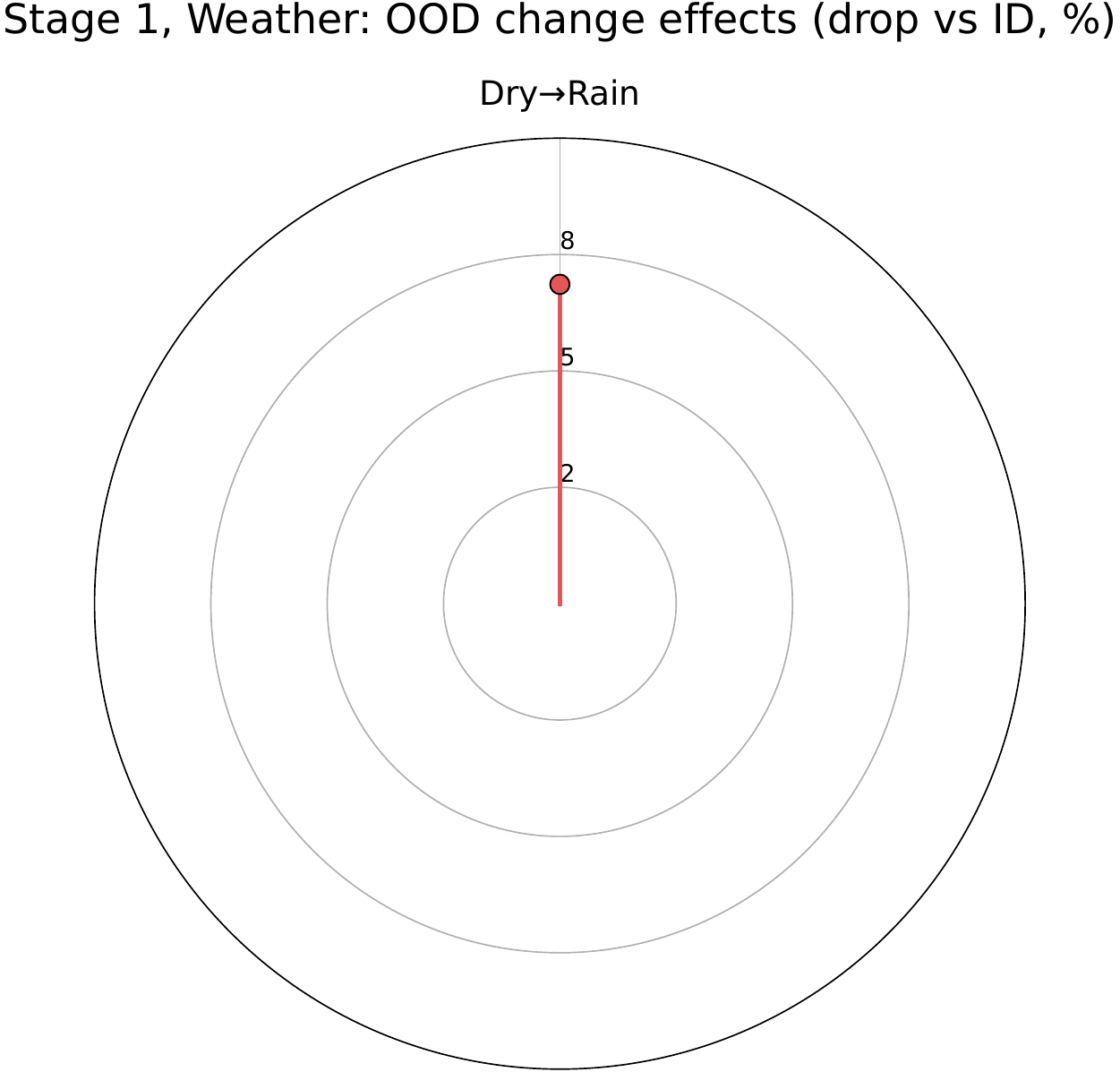}
  \caption{Weather}
\end{subfigure}\hfill
\begin{subfigure}[t]{0.32\textwidth}
  \centering
  \rule{0.98\linewidth}{0pt}
\end{subfigure}

\caption{Stage 1 themed star plots, single factor shifts.}
\label{fig:themed_star_stage1_all}
\vspace{-1.0ex}
\end{figure*}

This section provides themed decompositions of single factor shifts. Each subplot aggregates all shifts that share the same factor theme, highlighting which individual factors drive the largest mean degradation relative to the in distribution baseline.
See Fig.~\ref{fig:themed_star_stage1_all} for Stage 1 themed star plots summarizing single factor shifts, grouped by the shifted axis.

\subsection{Double factor themed star plots}
\begin{figure*}[!t]
\centering

\begin{subfigure}[t]{0.32\textwidth}
  \centering
  \includegraphics[height=0.21\textheight,keepaspectratio]{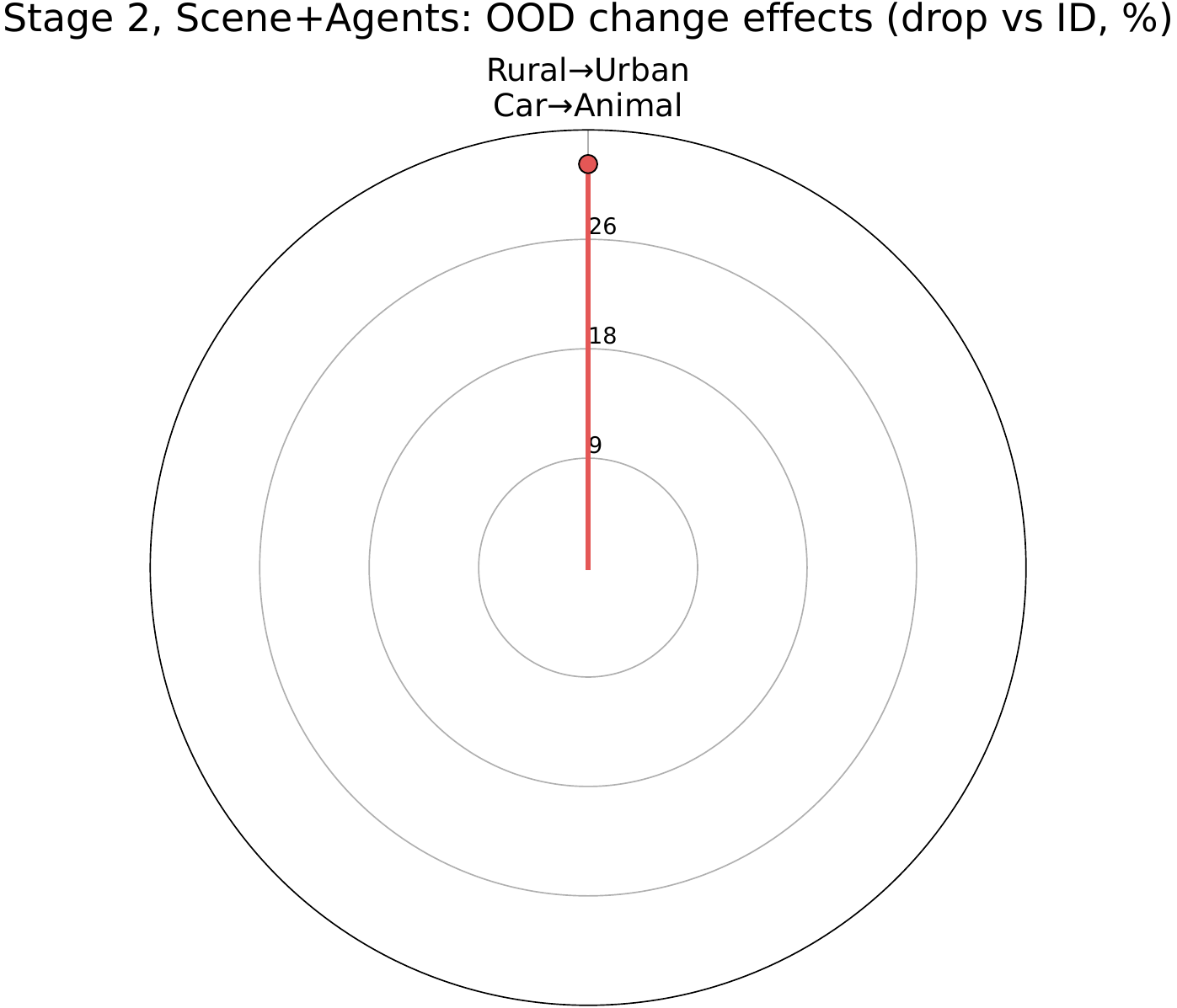}
  \caption{Scene, Agents}
\end{subfigure}\hfill
\begin{subfigure}[t]{0.32\textwidth}
  \centering
  \includegraphics[height=0.21\textheight,keepaspectratio]{Figures/Task3/Themes/Stage2/aggregate_stage2_theme_Scene_Season.pdf}
  \caption{Scene, Season}
\end{subfigure}\hfill
\begin{subfigure}[t]{0.32\textwidth}
  \centering
  \includegraphics[height=0.21\textheight,keepaspectratio]{Figures/Task3/Themes/Stage2/aggregate_stage2_theme_Scene_Time.pdf}
  \caption{Scene, Time}
\end{subfigure}

\vspace{0.8ex}

\begin{subfigure}[t]{0.32\textwidth}
  \centering
  \includegraphics[height=0.21\textheight,keepaspectratio]{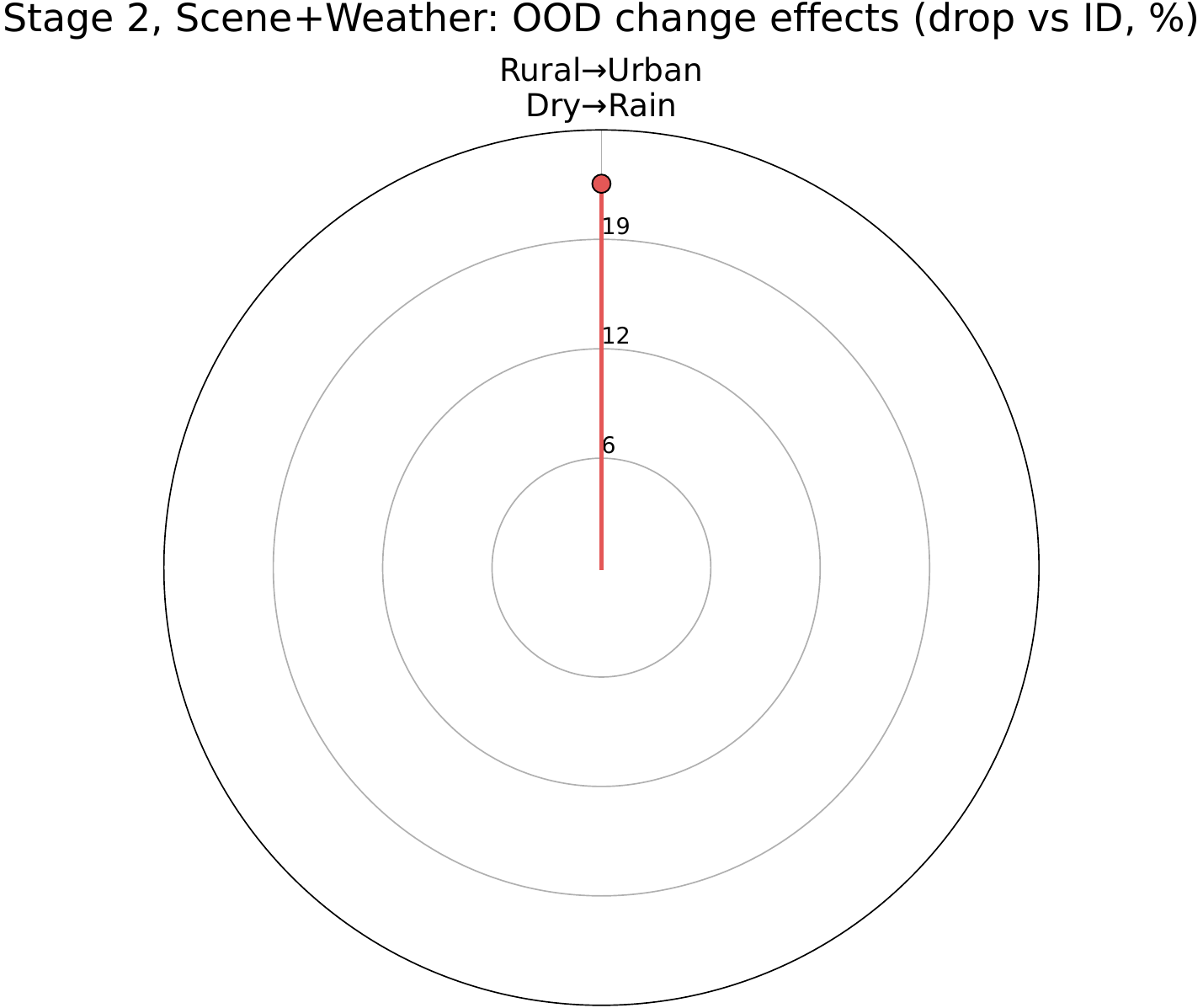}
  \caption{Scene, Weather}
\end{subfigure}\hfill
\begin{subfigure}[t]{0.32\textwidth}
  \centering
  \includegraphics[height=0.21\textheight,keepaspectratio]{Figures/Task3/Themes/Stage2/aggregate_stage2_theme_Season_Agents.pdf}
  \caption{Season, Agents}
\end{subfigure}\hfill
\begin{subfigure}[t]{0.32\textwidth}
  \centering
  \includegraphics[height=0.21\textheight,keepaspectratio]{Figures/Task3/Themes/Stage2/aggregate_stage2_theme_Season_Time.pdf}
  \caption{Season, Time}
\end{subfigure}

\vspace{0.8ex}

\begin{subfigure}[t]{0.32\textwidth}
  \centering
  \includegraphics[height=0.21\textheight,keepaspectratio]{Figures/Task3/Themes/Stage2/aggregate_stage2_theme_Season_Weather.pdf}
  \caption{Season, Weather}
\end{subfigure}\hfill
\begin{subfigure}[t]{0.32\textwidth}
  \centering
  \includegraphics[height=0.21\textheight,keepaspectratio]{Figures/Task3/Themes/Stage2/aggregate_stage2_theme_Time_Agents.pdf}
  \caption{Time, Agents}
\end{subfigure}\hfill
\begin{subfigure}[t]{0.32\textwidth}
  \centering
  \includegraphics[height=0.21\textheight,keepaspectratio]{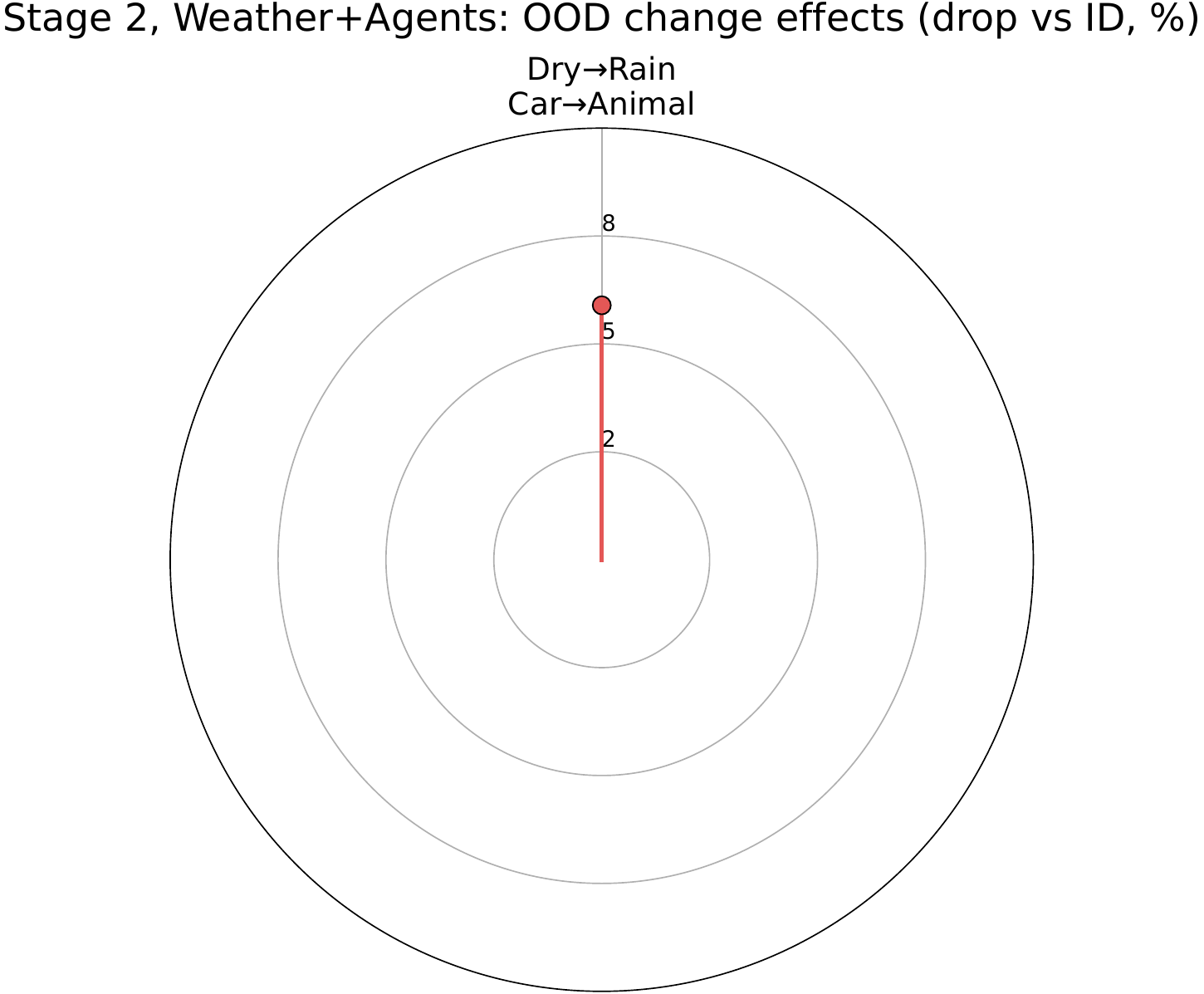}
  \caption{Weather, Agents}
\end{subfigure}

\caption{Stage 2 themed star plots, double factor shifts.}
\label{fig:themed_star_stage2_all}
\vspace{-1.0ex}
\end{figure*}

This section groups two factor shifts by their factor pair, enabling a direct view of which pairwise interactions amplify or mitigate degradation relative to single factor changes.
See Fig.~\ref{fig:themed_star_stage2_all} for Stage 2 themed star plots summarizing double factor shifts, grouped by the shifted factor pair.

\subsection{Triple factor themed star plots}
\begin{figure*}[!t]
\centering

\begin{subfigure}[t]{0.48\textwidth}
  \centering
  \includegraphics[height=0.22\textheight,keepaspectratio]{Figures/Task3/Themes/Stage3/aggregate_stage3_theme_Scene_Season_Time.pdf}
  \caption{Scene, Season, Time}
\end{subfigure}\hfill
\begin{subfigure}[t]{0.48\textwidth}
  \centering
  \includegraphics[height=0.22\textheight,keepaspectratio]{Figures/Task3/Themes/Stage3/aggregate_stage3_theme_Scene_Season_Weather.pdf}
  \caption{Scene, Season, Weather}
\end{subfigure}\hfill

\vspace{0.8ex}

\begin{subfigure}[t]{0.48\textwidth}
  \centering
  \includegraphics[height=0.22\textheight,keepaspectratio]{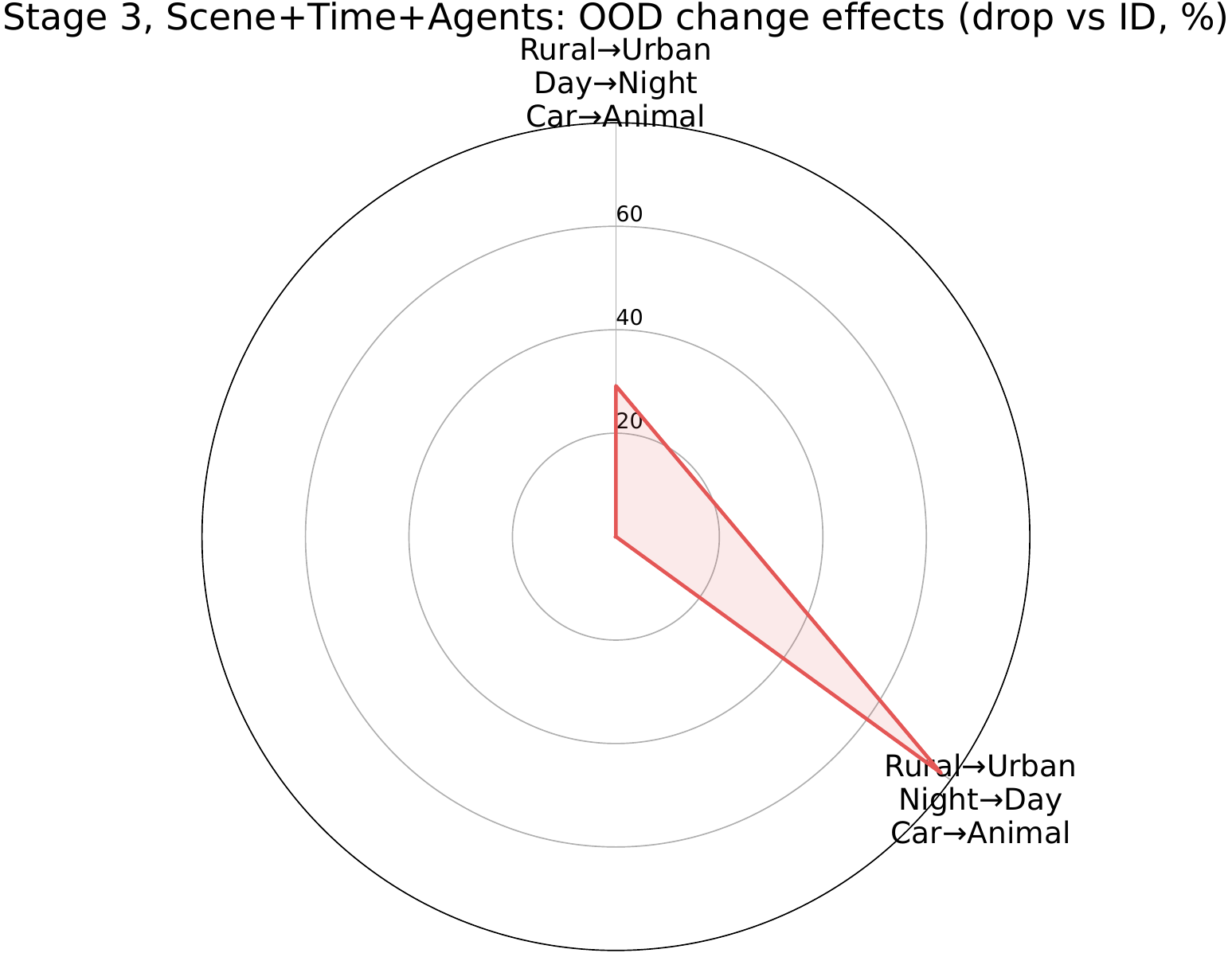}
  \caption{Scene, Time, Agents}
\end{subfigure}\hfill
\begin{subfigure}[t]{0.48\textwidth}
  \centering
  \includegraphics[height=0.22\textheight,keepaspectratio]{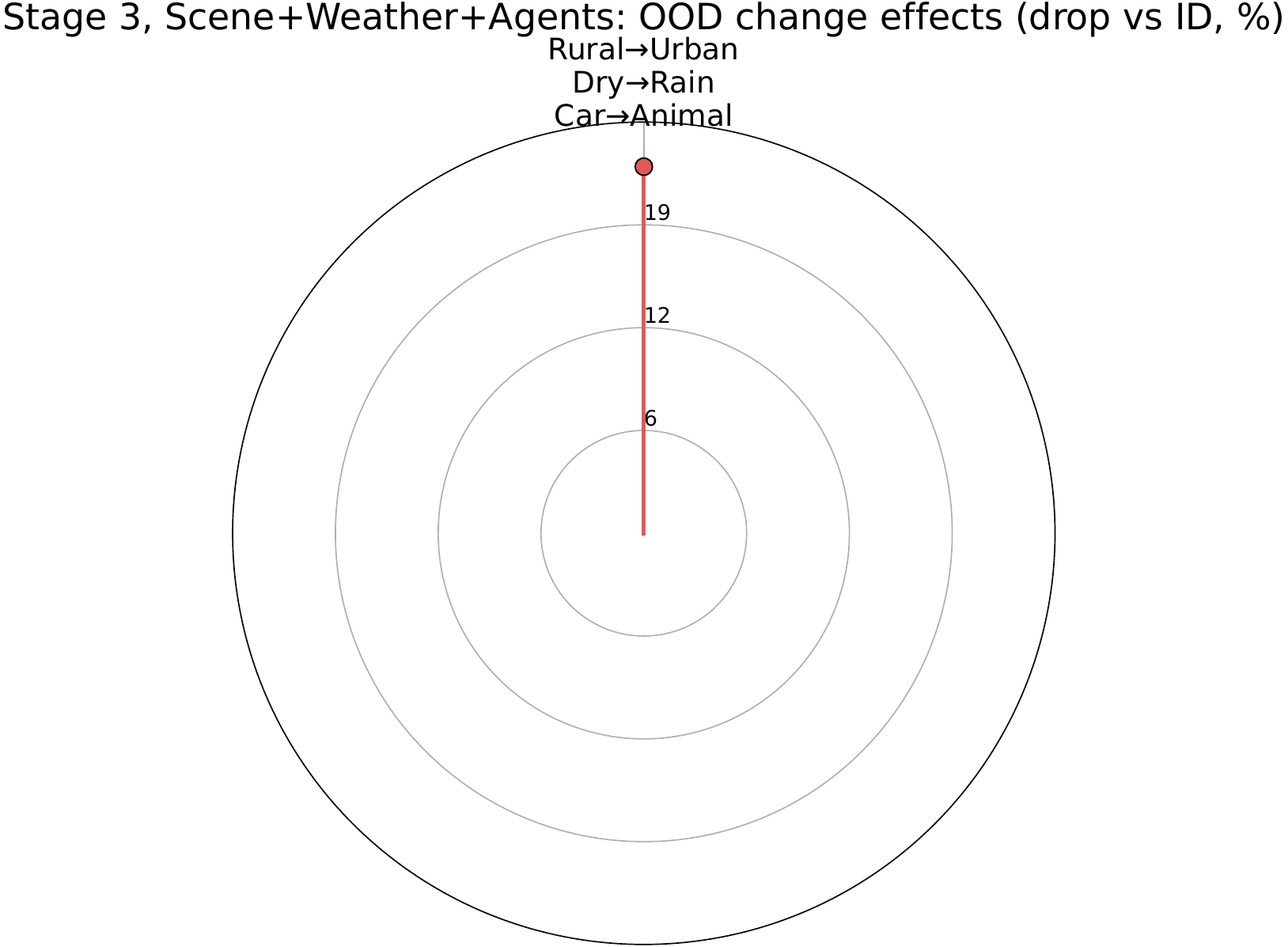}
  \caption{Scene, Weather, Agents}
\end{subfigure}

\vspace{0.8ex}

\begin{subfigure}[t]{0.48\textwidth}
  \centering
  \includegraphics[height=0.22\textheight,keepaspectratio]{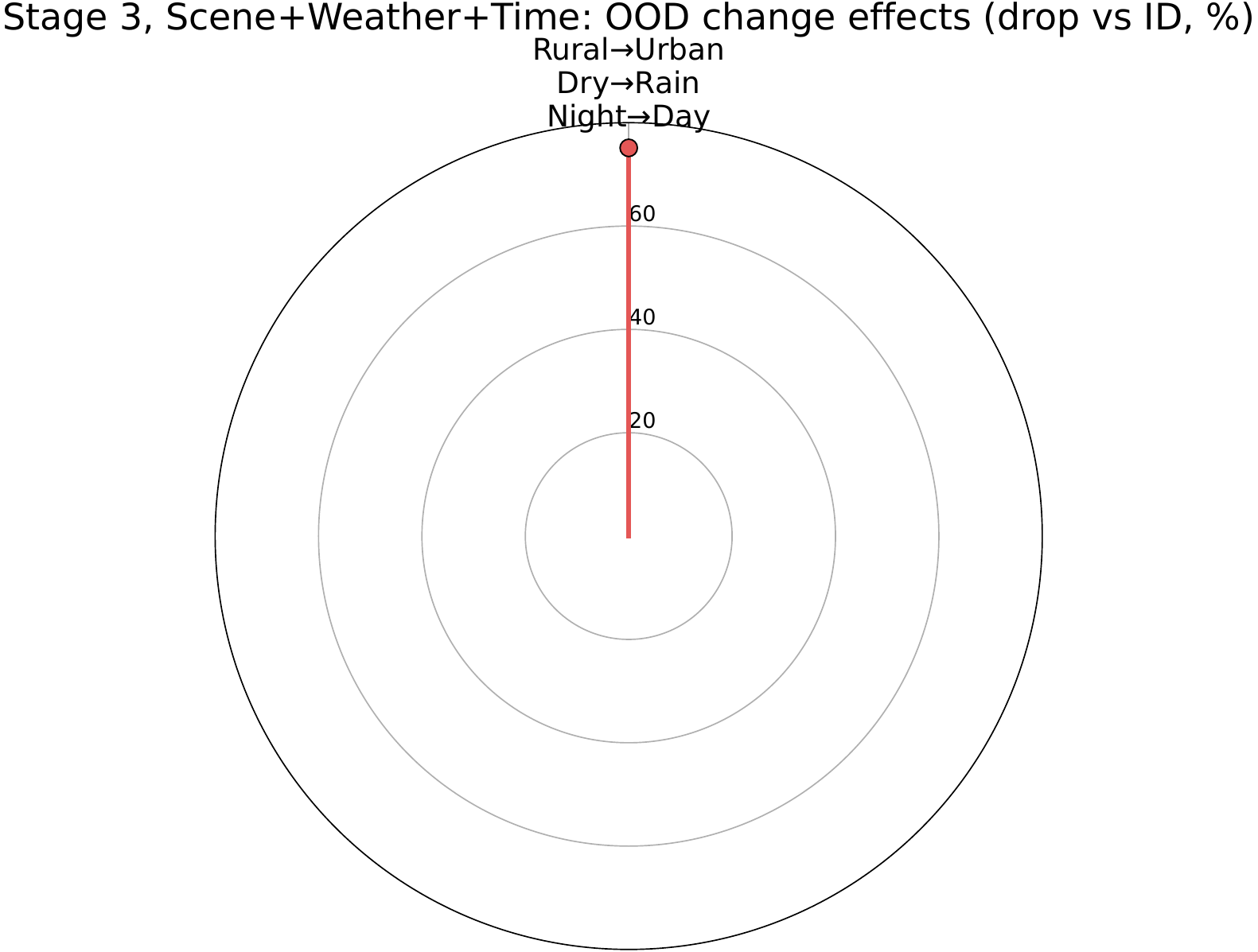}
  \caption{Scene, Weather, Time}
\end{subfigure}\hfill
\begin{subfigure}[t]{0.48\textwidth}
  \centering
  \includegraphics[height=0.22\textheight,keepaspectratio]{Figures/Task3/Themes/Stage3/aggregate_stage3_theme_Season_Time_Agents.pdf}
  \caption{Season, Time, Agents}
\end{subfigure}\hfill

\vspace{0.8ex}

\begin{subfigure}[t]{0.48\textwidth}
  \centering
  \includegraphics[height=0.22\textheight,keepaspectratio]{Figures/Task3/Themes/Stage3/aggregate_stage3_theme_Season_Weather_Agents.pdf}
  \caption{Season, Weather, Agents}
\end{subfigure}\hfill
\begin{subfigure}[t]{0.48\textwidth}
  \centering
  \includegraphics[height=0.22\textheight,keepaspectratio]{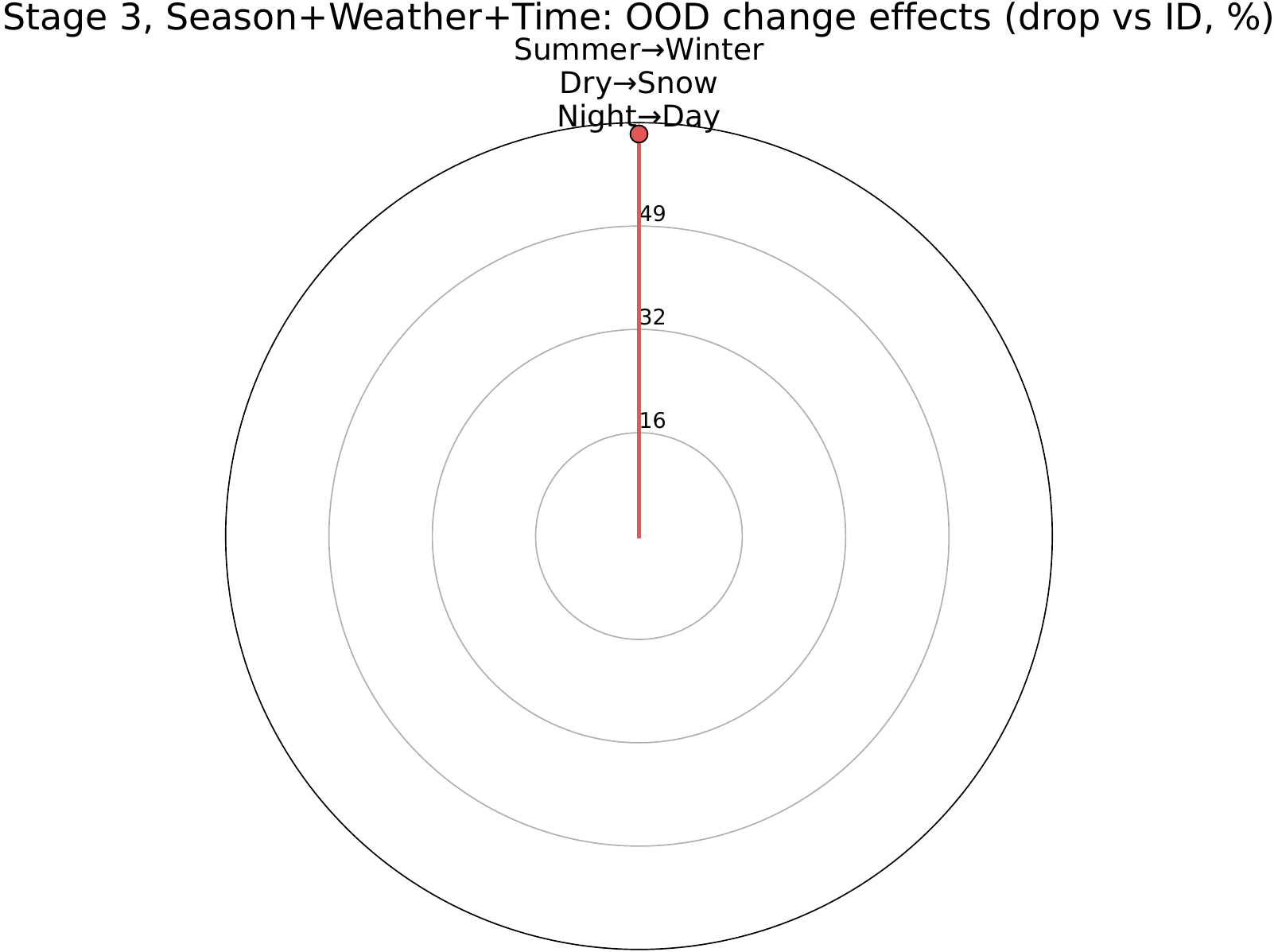}
  \caption{Season, Weather, Time}
\end{subfigure}

\caption{Stage 3 themed star plots, triple factor shifts.}
\label{fig:themed_star_stage3_all}
\vspace{-1.0ex}
\end{figure*}

This section aggregates three factor shifts by their factor triple, exposing which combinations compound most strongly and which remain comparatively stable under simultaneous changes.
See Fig.~\ref{fig:themed_star_stage3_all} for Stage 3 themed star plots summarizing triple factor shifts, grouped by the shifted factor triple.

\end{document}